\documentclass[sn-chicago]{sn-jnl}

\usepackage{graphicx}%
\usepackage{multirow}%
\usepackage{amsmath,amssymb,amsfonts}%
\usepackage{amsthm}%
\usepackage{mathrsfs}%
\usepackage[title]{appendix}%
\usepackage{xcolor}%
\usepackage{textcomp}%
\usepackage{manyfoot}%
\usepackage{booktabs}%
\usepackage{algorithmicx}%
\usepackage{algpseudocode}%
\usepackage{listings}%
\usepackage{longtable}%
\usepackage{natbib}
\usepackage[ruled]{algorithm2e}
\usepackage{hyperref}
\hypersetup{hyperfigures,breaklinks,colorlinks,linkcolor=blue,citecolor=blue,urlcolor=blue}

\makeatletter

\theoremstyle{thmstyleone}%
\theoremstyle{thmstyletwo}%

\theoremstyle{thmstylethree}%

\patchcmd{\NAT@citex}
  {\@citea\NAT@hyper@{%
     \NAT@nmfmt{\NAT@nm}%
     \hyper@natlinkbreak{\NAT@aysep\NAT@spacechar}{\@citeb\@extra@b@citeb}%
     \NAT@date}}
  {\@citea\NAT@nmfmt{\NAT@nm}%
   \NAT@aysep\NAT@spacechar\NAT@hyper@{\NAT@date}}{}{}

\patchcmd{\NAT@citex}
  {\@citea\NAT@hyper@{%
     \NAT@nmfmt{\NAT@nm}%
     \hyper@natlinkbreak{\NAT@spacechar\NAT@@open\if*#1*\else#1\NAT@spacechar\fi}%
       {\@citeb\@extra@b@citeb}%
     \NAT@date}}
  {\@citea\NAT@nmfmt{\NAT@nm}%
   \NAT@spacechar\NAT@@open\if*#1*\else#1\NAT@spacechar\fi\NAT@hyper@{\NAT@date}}
  {}{}

\makeatother

\begin{document}

\title[Article Title]{Review of Cloud Service Composition for Intelligent Manufacturing}


\author[1,2]{\fnm{Cuixia} \sur{Li}}\email{qyliying@126.com}
\equalcont{These authors contributed equally to this work.}

\author[1]{\fnm{Liqiang} \sur{Liu}}\email{liulq127@163.com}
\equalcont{These authors contributed equally to this work.}

\author*[2,3]{\fnm{Li} \sur{Shi}}\email{lcxxcl@zzu.edu.cn}
\equalcont{These authors contributed equally to this work.}

\affil[1]{\orgdiv{School of Cyber Science and Engineering}, \orgname{Zhengzhou University}, \orgaddress{\city{Zhengzhou}, \postcode{450001}, \country{China}}}

\affil[2]{\orgdiv{School of Electrical and Information Engineering}, \orgname{Zhengzhou University}, \orgaddress{\city{Zhengzhou}, \postcode{450001}, \country{China}}}

\affil*[3]{\orgdiv{Department of Automation}, \orgname{Tsinghua University}, \orgaddress{\city{Beijing}, \postcode{100084}, \country{China}}}


\abstract{Intelligent manufacturing is a new model that uses advanced technologies such as the Internet of Things, big data, and artificial intelligence to improve the efficiency and quality of manufacturing production. As an important support to promote the transformation and upgrading of the manufacturing industry, cloud service optimization has received the attention of researchers. In recent years, remarkable research results have been achieved in this field. For the sustainability of intelligent manufacturing platforms, in this paper we summarize the process of cloud service optimization for intelligent manufacturing. Further, to address the problems of dispersed optimization indicators and nonuniform/unstandardized definitions in the existing research, 11 optimization indicators that take into account three-party participant subjects are defined from the urgent requirements of the sustainable development of intelligent manufacturing platforms. Next, service optimization algorithms are classified into two categories, heuristic and reinforcement learning. After comparing the two categories, the current key techniques of service optimization are targeted. Finally, research hotspots and future research trends of service optimization are summarized.}




\maketitle

\section{Introduction}\label{sec1}
Intelligent manufacturing employs advanced information technology for achieving intelligence, automation, and digitalization in the manufacturing process \citep{num1}. It involves real-time data acquisition, analysis, and processing, along with the application of intelligent algorithms and control systems. These aid companies in optimizing resource allocation, enhancing production quality, lowering production costs, expediting product innovation and delivery, and fostering sustainable development. Intelligent manufacturing integrates cloud computing \citep{num2}, the Internet of Things (IoT) \citep{num3}, artificial intelligence (AI) \citep{num4}, big data \citep{num5}, and other emerging technologies. This integration enables the sharing of manufacturing resources and capabilities globally, forming a comprehensive intelligent manufacturing cloud service platform \citep{num6}. In an intelligent manufacturing cloud service platform, resource service providers release virtual manufacturing resource services. These services are centralized under the management of the cloud platform, forming cloud services \citep{num7}. Enterprise customers submit manufacturing tasks to the intelligent manufacturing platform. The platform decomposes tasks based on complexity and selects suitable cloud services from the candidate pool. These services are combined into a chain of cloud service composition to fulfill user manufacturing tasks \citep{num8}.

Cloud service optimization involves using intelligent algorithms to select the optimal service composition from numerous cloud services based on specific requirements, conditions, and constraints. The goal is to meet user needs, enhance system performance, reduce costs, and improve user experience \citep{num9}. This process includes operations such as selecting, combining, configuring, and scheduling cloud services to deliver optimal solutions. As a crucial component of intelligent manufacturing, cloud service optimization offers technical support by intelligently selecting and configuring diverse cloud service resources. It enables enterprises to maximize the use of cloud services for intelligent manufacturing, thereby enhancing production efficiency and product quality.

Intelligent Manufacturing Cloud Service Composition and Optimal Selection (IMfg-CSCOS) faces a series of complex issues, including cloud service providers, service models, optimization of resource service quality, etc. It requires careful selection of cloud services that meet customer needs. Additionally, factors such as service performance, reliability, security, and data privacy need to be thoroughly considered in the selection process. Meanwhile, research on cloud service optimization for intelligent manufacturing not only concerns the benefits of enterprises but also influences the competitiveness of the manufacturing industry in the market. Efficient cloud service composition solutions can substantially reduce customer costs and enhance satisfaction. However, inappropriate service composition solutions may affect customer satisfaction and limit the development of the intelligent manufacturing cloud platform. It is evident that IMfg-CSCOS is the key for traditional manufacturing to transition to intelligent manufacturing.

The goal of this review is to summarize IMfg-CSCOS research, covering selected source literature, the classification and canonical definition of existing optimization indicators, the classification and introduction of optimization algorithms, research hotspots, and future trends. The rest of this paper is structured as follows: Section \ref{sec2} describes the IMfg-CSCOS process. Section \ref{sec3} details the optimization objectives of IMfg-CSCOS. Section \ref{sec4} details the optimization methods of IMfg-CSCOS and briefly describes the evaluation indices of the optimization method. Section \ref{sec5} summarizes current research hot topics. Section \ref{sec6} illustrates the practical application scenarios of CSCOS in intelligent manufacturing. Section \ref{sec7} discusses the current challenges of CSCOS in intelligent manufacturing and outlines future research directions.

\section{IMfg-CSCOS Process}\label{sec2}

Intelligent manufacturing, as a new model, centralizes dispersed manufacturing resources using internet technology. It utilizes IoT, AI, cloud computing, and other technologies to encapsulate dispersed manufacturing resources into modules. These modules are then placed onto an intelligent manufacturing cloud platform for unified management. IMfg-CSCOS involves three participants: service providers, service demanders, and intelligent manufacturing platform operators. Service providers contribute distributed resources, which are encapsulated into manufacturing services on the intelligent manufacturing platform. Service demanders submit manufacturing requirements to the intelligent manufacturing platform, which in turn conducts cloud service optimization for the service demander. Intelligent manufacturing tasks can be categorized as single function and multifunction based on complexity. Single-function tasks are accomplished with a single service, whereas multifunctional tasks require decomposition into subtasks. These subtasks are completed by combining multiple heterogeneous manufacturing resources.

Given that tasks in real-world applications are typically more intricate, this paper focuses on the more extensively studied multifunctional tasks. Multifunctional manufacturing tasks have four types of structures: sequential, parallel, selective, and circular \citep{num10}. Sequential structure implies a unidirectional relationship between subtasks, executed in a specific order. Parallel structure involves the simultaneous execution of multiple unconstrained subtasks. Selection structures denote the presence of substitutions between specific subtasks, with the probability of each subtask occurring \citep{num11}. Circular structures incorporate loops in the subtask execution, requiring iterative execution. Zhou and Yao \citeyearpar{num12} demonstrated that parallel, selective, and circular structures can be simplified or transformed into the sequential model.

\begin{figure}
	\centering
	\includegraphics[width=0.9\linewidth]{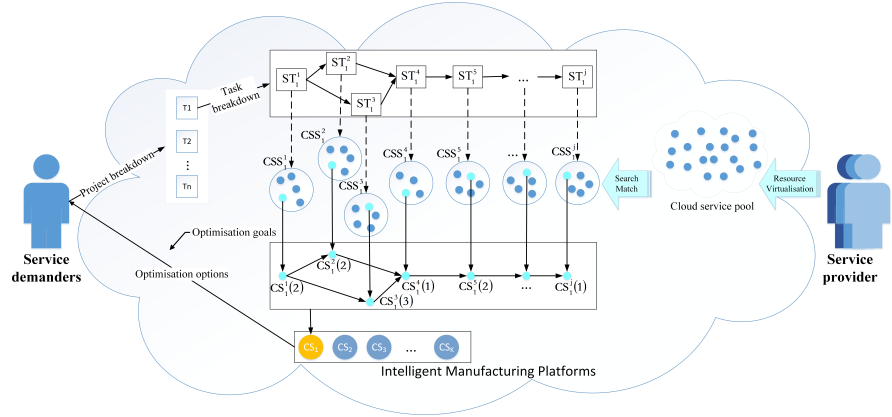}
	\caption{IMfg-CSCOS flow chart}
	\label{fig1}
\end{figure}

After determining the structure of the cloud service preference task, the intelligent manufacturing platform can optimize services based on this structure. Figure \ref{fig1} illustrates the process of IMfg-CSCOS. Using customer order \textit{O} as an example, the IMfg-CSCOS process can be described as follows: the customer order, \textit{O}, is decomposed into a series of tasks, \textit{T}. If \textit{T} is a multifunctional task, it continues to be decomposed into single-function subtasks, \textit{$ST_{i}$}. After that, services are searched based on the functional requirements of each subtask \textit{$ST_{i}$} to identify services that satisfy the subtasks’ quality of service (QoS) requirements, forming the candidate set of services for the subtasks. Finally, the optimal service composition is constructed by selecting services from the set of candidate services. The specific steps are as follows:

Step 1. \textit{Resource virtualization and servitization}: responsible for mapping shop floor physical resources to virtual resources.

Step 2. \textit{Task decomposition}: the user-submitted requirements are decomposed into a series of manufacturing tasks, which are further broken down into subtasks based on their functional characteristics.

Step 3. \textit{Search matching}: identify all manufacturing services that meet the subtask functionality and QoS requirements, and generate the corresponding set of candidate services.

Step 4. \textit{Manufacturing service composition and optimal selection}: select candidate services for subtasks based on the objective function, form a chain of service composition, and then execute an optimization algorithm to choose the optimal one.

Of these, Step 4 is critical and it directly affects the effectiveness of the service composition.

In practice, objective functions are categorized into positive (e.g., quality and reliability) and negative (e.g., time and cost), where the larger the value of the former, the higher the satisfaction of the service demanders, and vice versa for the latter. Therefore, the two types of indicators need to be normalized according to Equation (\ref{eq1}) and Equation (\ref{eq2}), respectively:

\begin{equation}\label{eq1}%
    {Nor}_{q}^{+} = \left\{ \begin{matrix}
{\frac{q - min~q}{max~q - min~q},~min~q \neq ~max~q} \\
{~~~~~~~~1,~~~~~~~~~min~q~ = ~max~q}
\end{matrix} \right.
\end{equation}

\begin{equation}\label{eq2}%
    {Nor}_{q}^{-} = \left\{ \begin{matrix}
{\frac{max~q - q}{max~q - min~q},~min~q \neq ~max~q} \\
{~~~~~~~~1,~~~~~~~~~min~q~ = ~max~q}
\end{matrix} \right.
\end{equation}

 where min \textit{q} represents the minimum value of the QoS attribute value in all combined chains, max \textit{q} denotes the corresponding maximum value, and \textit{q} denotes the QoS attribute value of the current service.
 
\section{IMfg-CSCOS Objectives}\label{sec3}

By reviewing the relevant literature and research outcomes, we have found that existing studies on IMfg-CSCOS predominantly concentrate on optimization objectives and diverse algorithms. These investigations have evolved from singular goals like cost and time optimization to multifaceted objectives such as reliability, quality, and reputation. Moreover, there has been a shift from the initial use of heuristic algorithms to the recent incorporation of advanced methods, notably reinforcement learning (RL). The research progress of IMfg-CSCOS is outlined below.

In establishing the canonical definition of optimization objectives, we analyzed 90 papers statistically. These articles were gathered through searches in the Web of Science and the China National Knowledge Infrastructure databases using keywords such as cloud manufacturing, service composition, and manufacturing services. After excluding 12 articles that did not specify a particular optimization objective, we were left with 78 papers. Addressing the sustainable development of cloud platforms, we identified 11 high-frequency optimization objectives. Based on the roles of participants in intelligent manufacturing activities, these objectives were categorized into three groups: those related to service demanders, service providers, and cloud platforms. The specific statistical results are presented in Table \ref{tab1}.

\begin{table}[h]
\caption{Statistical results of the optimization objectives of IMfg-CSCOS in the literature}\label{tab1}%
\begin{tabular}{@{}p{2cm}p{2cm}cp{6cm}@{}}
\toprule
     Participant& Optimization objectives & Quantity & Definition  \\
\midrule
Service Demander & Cost & 68 & The total cost of the process of completing the entire manufacturing task, including the cost of logistics and transportation. \\ 
& Time & 66 &   The time between the submission of a manufacturing task by a service demander and the completion of the task, including the time spent on logistics and transportation. \\ 
& Reliability & 28 &  The ability to successfully complete a manufacturing task within certain time and condition constraints.\\ 
& Quality & 28 &  The quality of the product produced during the manufacturing process.\\ 
& Reputation & 26 &  Reputation evaluation given by the user after the completion of the transaction.\\ 
& Availability & 16 &  Probability that a service is accessible in a certain period of time.\\ 
& Maintainability & 4 &  The probability that a cloud manufacturing service will be properly maintained and successfully complete its tasks in the event of an accident.\\
\midrule
Service Provider& Load Balancing & 5 &  Balanced distribution of manufacturing tasks among service providers.\\
\midrule
Cloud Platform& Resource Utilization Rate & 9 &  Usage efficiency of manufacturing resources in the service composition chain under manufacturing mode.\\ 
& Energy & 5 &  Total energy consumption in the service composition chain in manufacturing mode.\\ 
& Flexibility Indicator & 4 &  The ability to resist the occurrence of risk after encountering uncertainty.\\ 
\botrule
\end{tabular}
\end{table}

Referring to Table \ref{tab1}, the main objectives for cloud service optimization are cost and time, with secondary considerations given to reliability, quality, and reputation. Other factors such as availability, maintainability, load balancing, and energy come afterward.

In the current literature on IMfg-CSCOS, there are inconsistencies in the definition of the 11 optimization objectives mentioned above, hindering the progress of IMfg-CSCOS research. The varied terms and definitions used by different researchers introduce ambiguity to the same concepts, making the literature challenging to read and comprehend. Moreover, it diminishes the reproducibility and comparability of studies. To tackle this issue, we have provided a standardized definition for these 11 optimization objectives.

For the mentioned optimization objectives, let us consider an intelligent manufacturing cloud platform with D users ($d$ represents the $d$-th user) and P service providers ($p$ represents the $p$-th service provider) concurrently. Each user can request $T$ tasks from the platform, and each task comprises $J$ subtasks. Since the three structures (parallel, selective, and circular) can be simplified or transformed into the sequential model \citep{num12}, we only consider the sequential structure to define the above 11 optimization objectives using functions.

For analytical convenience, the following constraints should also be fulfilled:

(1) Tasks are independent of each other and share the same priority.

(2) Tasks are decomposed into subtasks based on requirements.

(3) Each subtask can be processed by at least one service.

(4) A service provider can offer different types of services, with each capable of performing a maximum of one subtask at a time.

(5) Subtasks cannot be interrupted until they are completed.

For descriptive purposes, notations were defined as shown in Table \ref{tab2}.

\begin{longtable}{@{}p{2cm}p{12cm}@{}}
\caption{Notation definitions}\label{tab2}\\
\hline
Notation & Definition  \\
\hline
$D$ & Cloud platform contains $D$ users  \\ 
$d$ & The $d$-th user of cloud platform  \\ 
$T$ & $T$ task requests per user to the cloud platform  \\ 
$i$ & The $i$-th task proposed by the user to the cloud platform  \\ 
$J$ & Each task contains $J$ subtasks  \\ 
$j$ & The $j$-th subtask of $T$  \\ 
$S$ & The cloud platform contains $S$ service providers  \\ 
$p$ & The $p$-th cloud service provider  \\ 
${K}_{p}$ & The $p$-th cloud service provider has  cloud service types  \\ 
$k$ & The $k$th cloud service of the $p$-th cloud service provider  \\ 
${st}_{ij}$ & The $j$-th subtask of the user's $i$-th task  \\ 
${X}_{ij}^{p,k}$ & Equals 1 when ${st}_{ij}$ is executed on ${S}_{p,k}$ and 0 otherwise  \\ 
${S}_{p,k}$ & The $k$th service of the $p$-th service provider  \\ 
${T}_{m}$ & Manufacturing consumption time committed by service providers  \\ 
${T}_{l}$ & Transportation time for manufacturing services  \\ 
${T}_{w}$ & Waiting time for these occupied services when the service arrives  \\ 
${WT}_{ij}^{pk}$ & Waiting time from $j$-th subtask to subsequent subtasks  \\ 
${LT}_{i(j,j+1)}^{pp'}$ & Transportation time from the $j$-th subtask to subsequent subtasks  \\ 
${CT}_{ij}^{p,k}$ & Manufacturing time of ${st}_{ij}$ on ${S}_{p,k}$  \\ 
${C}_{m}$ & Manufacturing costs of services  \\ 
${C}_{l}$ & Transportation costs of services  \\ 
${F}_{fu}^{p}$ & Diversity of service functions of the $p$-th service provider  \\ 
${F}_{ty}^{p}$ & Type of resources owned by the $p$-th service provider  \\ 
${F}_{co}^{p}$ & Number of cooperating enterprises of the $p$-th service provider  \\ 
${F}_{sa}^{p}$ & Number of spare resources available to the $p$-th service provider  \\
$SE$ & Energy consumed by cloud services manufacturing  \\ 
$LE$ & Energy consumed by transport in the manufacture of cloud services  \\ 
$FLr$ & The ability of service provider to handle changes in manufacturing resources  \\ 
${ma}_{ij}^{p,k}$ & Maintainability of ${st}_{ij}$ on ${S}_{p,k}$  \\ 
${N}_{am}^{p,k}$ & Number of times service ${S}_{p,k}$ has been successfully maintained in the event of an unexpected situation in the history of invocations  \\ 
${N}_{m}^{p,k}$ & Total number of times service ${S}_{p,k}$ has had an unexpected situation in its history of invocations  \\ 
${q}_{ij}^{p,k}$ & QoS for ${st}_{ij}$ on ${S}_{p,k}$ \\ 
${s}_{ij}^{p,k}$ & Service reputation for ${st}_{ij}$ on ${S}_{p,k}$  \\ 
${E}_{h}^{p,k}$ & Reputation assessment of ${S}_{p,k}$  \\ 
${N}_{e}$ & Total number of reputation evaluations  \\ 
$h$ & The $h$-th reputation evaluation of ${S}_{p,k}$  \\ 
${a}_{ij}^{p,k}$ & Service availability of ${st}_{ij}$ on ${S}_{p,k}$ \\ 
${N}_{ar}^{p,k}$ & Number of successful executions of ${S}_{p,k}$ in the most recent time interval ${t}_{total}^{p,k}$  \\ 
${N}_{r}^{p,k}$ & Total number of times service ${S}_{p,k}$ has been invoked in the most recent time interval ${t}_{total}^{p,k}$  \\ 
${r}_{ij}^{p,k}$ & Reliability of ${st}_{ij}$ over ${S}_{p,k}$  \\ 
${t}_{total}^{p,k}$ & Recent time interval for service provider ${S}_{p,k}$  \\ 
${t}_{avail}^{p,k}$ & Trouble-free time of service provider ${S}_{p,k}$ at ${t}_{total}^{p,k}$  \\ 
${C}_{c}$ & Service fees for cloud platforms  \\ 
${CC}_{ij}^{p,k}$ & Cost of ${st}_{ij}$ on ${S}_{p,k}$  \\ 
${LC}_{i(j,j+1)}^{pp'}$ & Transportation cost from the $j$-th subtask to subsequent subtasks  \\ 
${PT}_{ij}^{p,k}$ & Manufacturing spend of ${st}_{ij}$ on ${S}_{p,k}$  \\ 
${MSC}_{i}$ & The set of service providers selected for the chain of service compositions for the $i$-th task  \\ 
${MSN}_{i}$ & Number of service provider selected for the service composition chain of the $i$-th task  \\ 
${LE}_{i(j,j+1)}^{p,k}$ & Transportation energy consumption of subsequent subtasks of the $j$-th subtask  \\ 
${E}_{ij}^{p,k}$ & Energy consumed by ${st}_{ij}$ in manufacturing on ${S}_{p,k}$ \\ 

$FLt$ & The ability of service provider to handle changes in manufacturing tasks  \\ 
${Load}_{p}$ & Workload of the $p$-th service provider  \\ 
$STIME$ & The sum of actual available manufacturing time for all service providers  \\ 
\hline
\end{longtable}

Conforming to the definitions given in Table \ref{tab3}, a service optimization objective function that takes into account the three participating subjects can be defined, as shown in Equations (\ref{eq3})–(\ref{eq26}).

\begin{table}[h]
\caption{Statistical results of the optimization objectives of IMfg-CSCOS in the literature}\label{tab3}%
\begin{tabular}{@{}|p{2cm}|p{4cm}|p{2cm}|p{4cm}|@{}}
\toprule
     Function & Definition & Function & Definition  \\
\hline
$f_{1}$ & Cost & $f_{7}$ & Maintainability \\ 
\hline
$f_{2}$ & Time & $f_{8}$ & Load Balancing \\ 
\hline
$f_{3}$ & Reliability & $f_{9}$ & Resource Utilization Rate \\ 
\hline
$f_{4}$ & Quality & $f_{10}$ & Energy \\ 
\hline
$f_{5}$ & Reputation & $f_{11}$ & Flexibility Indicator \\ 
\hline
$f_{6}$ & Availability &  &  \\ 
\botrule
\end{tabular}
\end{table}

\begin{equation}\label{eq3}%
    f_{1} = min\left( C_{m} + C_{l} + C_{c} \right)
\end{equation}
where $C_{m}$ and $C_{l}$ are defined as in Equations (\ref{eq4}) and (\ref{eq5}):
\begin{equation}\label{eq4}%
C_{m} = {\sum\limits_{i = 1}^{T}\,}{\sum\limits_{j = 1}^{J}\,}{\sum\limits_{p = 1}^{S}\,}\,{\sum\limits_{k = 1}^{K_{p}}\,}{CC}_{ij}^{p,k} \times X_{ij}^{p,k}
\end{equation}
\begin{equation}\label{eq5}%
C_{l} = {\sum\limits_{i = 1}^{T}\,}{\sum\limits_{j = 1}^{J - 1}\,}\left( {{\sum\limits_{p = 1}^{S}\,}{\sum\limits_{\substack{p' = 1 \\ p \neq p'}}^{S}\,}\,{\sum\limits_{k = 1}^{K_{p}}\,}{\,{\sum\limits_{k' = 1}^{K_{p}}\,}X_{ij}^{p,k}} \times X_{i(j + 1)}^{p',k'} \times {LC}_{i(j,j + 1)}^{pp’}} \right)
\end{equation}
\begin{equation}\label{eq6}%
    f_{2} = min\left( T_{m} + T_{l} + T_{w} \right)
\end{equation}
where $T_{m}$, $T_{l}$, and $T_{w}$ are respectively defined in Equations (\ref{eq7})–(\ref{eq9}):
\begin{equation}\label{eq7}%
T_{m} = {\sum\limits_{i = 1}^{T}\,}{\sum\limits_{j = 1}^{J}\,}{\sum\limits_{p = 1}^{S}\,}\,{\sum\limits_{k = 1}^{K_{p}}\,}{CT}_{ij}^{p,k} \times X_{ij}^{p,k}
\end{equation}
\begin{equation}\label{eq8}%
T_{l} = {\sum\limits_{i = 1}^{T}\,}{\sum\limits_{j = 1}^{J - 1}\,}\left( {{\sum\limits_{p = 1}^{S}\,}{\sum\limits_{\substack{p' = 1 \\ p \neq p'}}^{S}\,}\,{\sum\limits_{k = 1}^{K_{p}}\,}{\,{\sum\limits_{k' = 1}^{K_{p}}\,}X_{ij}^{p,k}} \times X_{i(j + 1)}^{p',k'} \times {LT}_{i(j,j + 1)}^{pp’}} \right)
\end{equation}
\begin{equation}\label{eq9}%
T_{w} = {\sum\limits_{i = 1}^{T}\,}{\sum\limits_{j = 1}^{J}\,}{\sum\limits_{p = 1}^{S}\,}\,{\sum\limits_{k = 1}^{K_{p}}\,}X_{ij}^{p,k} \times {WT}_{ij}^{p,k}
\end{equation}
\begin{equation}\label{eq10}%
f_{3} = max\left( {{\sum\limits_{i = 1}^{T}\,}\,\,\left( {{\sum\limits_{j = 1}^{J}\,}\,\left( {{\sum\limits_{p = 1}^{S}\,}\,{\sum\limits_{k = 1}^{K_{p}}\,}r_{ij}^{p,k} \times X_{ij}^{p,k}} \right)} \right)} \right)
\end{equation}
where $r_{ij}^{p,k}$ are defined in Equation (\ref{eq11}):
\begin{equation}\label{eq11}%
r_{ij}^{p,k} = \frac{N_{ar}^{p,k}}{N_{r}^{p,k}}
\end{equation}
\begin{equation}\label{eq12}%
f_{4} = max\left( {{\sum\limits_{i = 1}^{T}\,}\,\,\left( {{\prod\limits_{j = 1}^{J}\,}\,\left( {{\sum\limits_{p = 1}^{S}\,}\,{\sum\limits_{k = 1}^{K_{p}}\,}\,q_{ij}^{p,k} \times X_{ij}^{p,k}} \right)} \right)} \right)
\end{equation}
\begin{equation}\label{eq13}%
f_{5} = max{\sum\limits_{i = 1}^{T}\,}{\sum\limits_{j = 1}^{J}\,}\,\left( {{\sum\limits_{p = 1}^{S}\,}\,{\sum\limits_{k = 1}^{K_{p}}\,}\,s_{ij}^{p,k} \times X_{ij}^{p,k}} \right)
\end{equation}
where $s_{ij}^{p,k}$ are defined in Equation (\ref{eq14}):
\begin{equation}\label{eq14}%
s_{ij}^{p,k} = \frac{\sum\limits_{h = 1}^{N_{e}}E_{h}^{p,k}}{N_{e}}
\end{equation}
\begin{equation}\label{eq15}%
f_{6} = max{\sum\limits_{i = 1}^{T}\,}\,{\sum\limits_{j = 1}^{J}\,}\,\left( {{\sum\limits_{p = 1}^{S}\,}\,{\sum\limits_{k = 1}^{K_{p}}\,}\,a_{ij}^{p,k} \times X_{ij}^{p,k}} \right)
\end{equation}
where $a_{ij}^{p,k}$ are defined in Equation (\ref{eq16}):
\begin{equation}\label{eq16}%
a_{ij}^{m,k} = \frac{t_{avail}^{p,k}}{t_{total}^{p,k}}
\end{equation}
\begin{equation}\label{eq17}%
f_{7} = max{\sum\limits_{i = 1}^{T}\,}\,{\sum\limits_{j = 1}^{J}\,}\,\left( {{\sum\limits_{p = 1}^{S}\,}\,{\sum\limits_{k = 1}^{K_{p}}\,}\,{ma}_{ij}^{p,k} \times X_{ij}^{p,k}} \right)
\end{equation}
where ${ma}_{ij}^{p,k}$ are defined in Equation (\ref{eq18}):
\begin{equation}\label{eq18}%
{ma}_{ij}^{p,k} = \frac{N_{am}^{p,k}}{N_{m}^{p,k}}
\end{equation}
\begin{equation}\label{eq19}%
f_{8} = min~\sigma\,\left( {Load}_{p} \right),p \in S
\end{equation}
where ${Load}_{p}$ are defined in Equation (\ref{eq20}):
\begin{equation}\label{eq20}%
{Load}_{p} = {\sum\limits_{i = 1}^{T}\,}{\sum\limits_{j = 1}^{J}\,}{\sum\limits_{k = 1}^{K_{p}}\,}CT_{ij}^{p,k} \times X_{ij}^{p,k},p \in \left\{ 1,2,...,S \right\}
\end{equation}
\begin{equation}\label{eq21}%
f_{9} = max\left( \frac{{\sum\limits_{i = 1}^{T}\,}{\sum\limits_{j = 1}^{J}\,}{\sum\limits_{p = 1}^{S}\,}\,{\sum\limits_{k = 1}^{K_{p}}\,}{CT}_{ij}^{p,k} \times X_{ij}^{p,k}}{STIME} \right)
\end{equation}
\begin{equation}\label{eq22}%
f_{10} = min(SE + LE)
\end{equation}
where $SE$ and $LE$ are respectively defined in Equations (\ref{eq23}) and (\ref{eq24}):
\begin{equation}\label{eq23}%
SE = {\sum\limits_{i = 1}^{T}\,}{\sum\limits_{j = 1}^{J}\,}{\sum\limits_{p = 1}^{S}\,}\,{\sum\limits_{k = 1}^{K_{p}}\,}E_{ij}^{p,k} \times X_{ij}^{p,k}
\end{equation}
\begin{equation}\label{eq24}%
LE = {\sum\limits_{i = 1}^{T}\,}{\sum\limits_{j = 1}^{J - 1}\,}\left( {{\sum\limits_{p = 1}^{S}\,}{\sum\limits_{\substack{p' = 1 \\ p \neq p'}}^{S}\,}\,{\sum\limits_{k = 1}^{K_{p}}\,}{\,{\sum\limits_{k' = 1}^{K_{p}}\,}X_{ij}^{p,k}} \times X_{i(j + 1)}^{p',k'} \times {LE}_{i(j,j + 1)}^{pp’}} \right)
\end{equation}
\begin{equation}\label{eq25}%
f_{11} = max(FLt + FLr)
\end{equation}
where $FLt$ and $FLr$ are defined, respectively, in Equations (\ref{eq26}) and (\ref{eq27}):
\begin{equation}\label{eq26}%
FLt = {\sum\limits_{i = 1}^{T}{\frac{\sum\limits_{p = 1,p \in {MCS}_{i}}^{{MSN}_{i}}{F_{fu}^{p}\,}}{{MSN}_{i}} + \frac{\sum\limits_{p = 1,p \in {MCS}_{i}}^{{MSN}_{i}}{F_{ty}^{p}\,}}{{MSN}_{i}} + \frac{\sum\limits_{p = 1,p \in {MCS}_{i}}^{{MSN}_{i}}{F_{co}^{p}\,}}{{MSN}_{i}}}}
\end{equation}
\begin{equation}\label{eq27}%
FLr = {\sum\limits_{i = 1}^{T}{\frac{\sum\limits_{p = 1,p \in {MCS}_{i}}^{{MSN}_{i}}{F_{sa}^{p}\,}}{{MSN}_{i}} + \frac{\sum\limits_{p = 1,p \in {MCS}_{i}}^{{MSN}_{i}}{F_{co}^{p}\,}}{{MSN}_{i}}}}
\end{equation}

Based on the involved participant subjects, existing studies can be categorized into three types: focusing on a single participant subject, involving two participant subjects, and addressing a three-party participant subject.

(1) \textit{For individual participants}. In IMfg-CSCOS studies, the service demander typically plays a central role in decision making, and research oriented towards individual participants aligns with the service demander’s interests. Besides the primary optimization goals of cost and time, customers also prioritize reliability, quality, and reputation. Zhou and Yao \citeyearpar{num13} expanded considerations to include reputation and reliability, while Akbaripour et al. \citeyearpar{num14} additionally factored in quality. Availability, defined as the probability of normal operation of a manufacturing service \citep{num15}, is a critical metric. For example, Zhou et al. \citeyearpar{num16} incorporated it to meet customer manufacturing needs. Maintainability refers to a manufacturing service’s ability to continue tasks and resolve anomalies under abnormal conditions; e.g., Tang et al. \citeyearpar{num17} optimized for these objectives to enhance the manufacturing service chain’s resilience to unexpected situations.

(2) \textit{Considering two participants}. When focusing solely on service optimization from the customer’s viewpoint, there is a risk of neglecting the interests of the intelligent manufacturing platform and service providers. Neglecting the interests of the intelligent manufacturing platform may lead to inadequate management, allowing false services to disrupt the market \citep{num18}. Similarly, overlooking the interests of service providers can impact their motivation, degrade the user’s experience, and even result in providers withdrawing from the manufacturing platform in severe cases \citep{num19}. Considering these factors, Hu et al. \citeyearpar{num20} enhanced service provider motivation by optimizing load balancing to achieve a balanced benefit distribution while addressing customer interests. Einollah Jafarnejad Ghomi and Qader \citeyearpar{num21} minimized time and cost while also coordinating benefits among service providers through load balancing optimization. Tao et al. \citeyearpar{num22} increased cloud platform revenue by minimizing energy consumption, and Zhu et al. \citeyearpar{num23} reduced resource consumption by maximizing resource utilization for cloud platforms. Wei et al. \citeyearpar{num24} enhanced the flexibility of cloud platforms by optimizing flexibility metrics.

(3)\textit{Considering three participants}. In a three-party situation, if only the interests of two parties are taken into account, it will inevitably lead to the dissatisfaction of one of the remaining participants. Comprehensive consideration of all three participants involved in the intelligent manufacturing process within the scope of the optimization goal promotes the motivation of all participants and creates a favorable intelligent manufacturing environment. Wang et al. \citeyearpar{num25} constructed an eight-objective optimization model that ensures customer benefits by minimizing the service time and maximizing service quality and reliability, which increases the profitability of the cloud platform by reducing the total cost and energy consumption and improves the profitability of the service provider by reducing load balancing and decreasing service idle time. Lim et al. \citeyearpar{num19} considered the interests of all participants in intelligent manufacturing by minimizing the time and cost and maximizing the QoS to satisfy the user’s needs. They also focused on maximizing resource utilization to ensure that the benefits of the cloud platform are maximized, and maximizing the economic benefits of the enterprise to enable the service provider to capture more manufacturing tasks and profits. Chen et al. \citeyearpar{num26} ensured user benefits by minimizing the time and cost while ensuring service provider motivation and cloud platform benefits with load balancing and maximizing resource utilization. Xiong et al. \citeyearpar{num27} considered the interests of three stakeholders, analyzed optimization objectives, and identified potential conflict situations that could impact the interests of multiple stakeholders in the service combination process. They eventually proposed an adaptive adjustment model based on the interests of these stakeholders.

\section{Service Optimization Methods}\label{sec4}

The IMfg-CSCOS problem fundamentally constitutes a composition optimization challenge. An intelligent manufacturing platform harbors a substantial and irregular array of cloud services, embodying a quintessential nondeterministic polynomial time–hardness problem. In response to such intricacies, intelligent algorithms have emerged as prevalent tools for effective resolution in academic contexts \citep{num28}. In the initial phases of research, heuristic algorithms were extensively applied in the domain of IMfg-CSCOS owing to their straightforward principles, applicability, and broad relevance. As machine learning and RL have progressed, intelligent algorithms are now increasingly employed to address optimization challenges in cloud services. The subsequent sections will delineate each of the three categories of IMfg-CSCOS-solving algorithms along with their notable representatives.

\subsection{Heuristic Optimization Algorithms}\label{subsec4.1}

Heuristic algorithms represent computational approaches for addressing intricate optimization problems, providing practical solutions within a reasonable temporal and spatial framework. In contrast to conventional mathematical methods, heuristic algorithms prioritize exploration within the domain of approximate solutions, facilitating the swift identification of improved outcomes.

\begin{figure}
	\centering
	\includegraphics[width=0.5\linewidth, height=0.6\textheight]{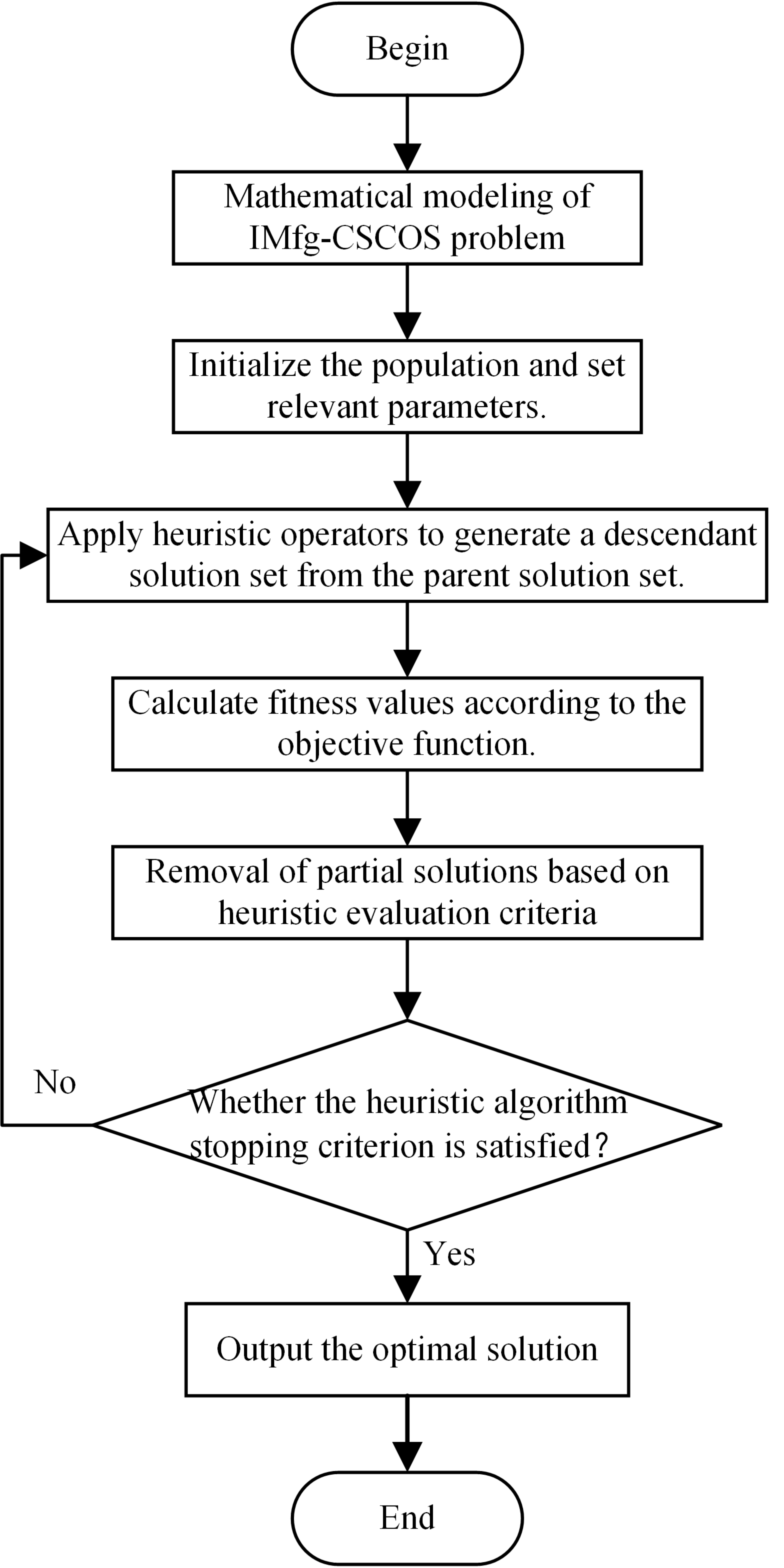}
	\caption{Heuristic algorithm flow chart}
	\label{fig2}
\end{figure}

The heuristic algorithm workflow is illustrated in Figure \ref{fig2}. Initially, it is essential to define the solution, objective function, and the corresponding value of the problem. Subsequently, the problem is modeled, as outlined in Algorithm \ref{alg1}.

\begin{algorithm}[H]
	\caption{Heuristic algorithm for solving the IMfg-CSCOS problem}\label{alg1}

	Mathematical modeling of IMfg-CSCOS problem;\\
	Define the objective function based on the objectives of the IMfg-CSCOS problem;\\
	Initialize the population and set relevant parameters;\\
	\While{Termination conditions}
	{
		According to the heuristic algorithm the operator generates a batch of 
		new solutions from the set of solutions, calculates the objective function value 
		and merges it with the previous set of solutions;\\
		According to the heuristic evaluation criterion, a portion of the solutions that 
		do not conform to the requirements are removed;
	}
	Return the best solution
\end{algorithm}


In the current stage of research, heuristic algorithms are predominantly characterized by imitating natural biological processes. Widely employed heuristic algorithms encompass genetic algorithm (GA) \citep{num29}, particle swarm optimization (PSO) \citep{num30} algorithms, and chaotic optimization algorithm. Additionally, there is utilization of the whale optimization algorithm (WOA) \citep{num31}, ant colony optimization (ACO) \citep{num32} algorithms, artificial bee colony(ABC) \citep{num33} algorithm, simulated annealing (SA) \citep{num34} algorithm, and others. The classification of heuristic algorithms is based on problem-solving approaches, distinguishing between swarm algorithms, evolutionary algorithms (EAs), physical optimization algorithms, and hybrid algorithms \citep{num35}.

Swarm algorithms \citep{num36} draw inspiration from the collective and cooperative behaviors observed in various species, such as ants, birds, and bees, where individuals work together in groups to find and collect food. Prominent swarm heuristic algorithms encompass ACO, PSO, ABC, and the emerging WOA. For instance, the ACO algorithm emulates the behavior of ants that release pheromones along their foraging routes. The concentration of pheromones serves as an indicator of path distance, where higher concentrations correlate with shorter paths. Cao et al. \citeyearpar{num37} introduced an enhanced ACO algorithm that integrates a selection mechanism into the standard ACO to address cloud service optimization problems. In this modification, ants choose their routes based on transfer probabilities; however, the selected route may be invalid. To address this, a selection mechanism is incorporated to update the pheromone concentration. This mechanism ensures that only ants with valid routes are retained, while those with invalid routes are eliminated. The PSO algorithm, introduced by Eberhart and Kennedy \citeyearpar{num38}, is a collaborative search algorithm inspired by the foraging behavior of bird flocks. The fundamental concept of this algorithm involves individuals within a group sharing information, facilitating the movement of the entire group within the problem-solving space. This collective movement induces an evolutionary process, transitioning from disorder to order, ultimately leading to the identification of the optimal solution for the problem at hand. Xie et al. \citeyearpar{num39} introduced a two-stage approach: in the initial stage, K-means clustering is employed to reduce the solution space and enhance the quality of candidate services, emphasizing service stability; the subsequent stage employs a chaotic Gaussian multi-objective PSO algorithm to identify the optimal service composition solution. Chaotic sequences are employed for population initialization and promoting diversity, while Gaussian perturbation operators are introduced for population updates to mitigate the risk of converging to local optima. Drawing inspiration from the collective hunting behavior of humpback whales in nature, Mirjalili and Lewis \citeyearpar{num31} introduced a novel swarm intelligence optimization algorithm called WOA in 2016. While WOA is designed for solving single-objective continuous problems and may not be directly applicable to address cloud service optimization problems, Wang et al. \citeyearpar{num40} proposed enhancement strategies such as dynamic coding and waiting cost optimization based on WOA, resulting in the development of the adaptive multi-objective WOA.

EAs are commonly employed to address problems characterized by a vast solution space \citep{num41}. Modeled after the principles of biological evolution—such as genetic crossover, selection, and mutation—an EA generates an initial set of solutions and iteratively refines them to yield optimal solutions. Among the most prevalent and representative EAs are GA  and non-dominated sorting genetic algorithm (NSGA) \citep{num42}. For instance, Jiang et al. \citeyearpar{num43} devised an enhanced variable-length encoding GA, inspired by Charles Darwin’s biological evolution theory, to tackle uncertainties and dynamism in the cloud service optimization process. The uncertain aspect of the service composition process is represented as an abstract service, and dynamical changes in the process model are accommodated by utilizing progressively increasing gene fragments. This approach mitigates the impact of dynamical changes to a manageable level. Furthermore, crossover and mutation algorithms, tailored for nonlinearly varying structures and individuals with incremental service composition, are designed to address the cloud service preference problem. NSGA-II, which is discussed further below, is an enhanced version of the NSGA introduced by Deb et al. \citeyearpar{num44}. Lim et al. \citeyearpar{num19} presented a service composition model based on a three-layer programming model, which proves challenging for NSGA-II due to the complexity of the model’s optimal solution. To address this, an improved NSGA-II algorithm with advancements and inheritance (a-i-NSGA-II) was proposed, which improves the coding method and group screening mechanism of the NSGA-II algorithm to mitigate local optimization issues. The crowding distance is solved by considering the objective value of the next level in advance, establishing stronger correlations between levels. The input solution of the lower level inherits the output solution of the upper level, enhancing the reliability of the solution for the three-layer model. Finally, there are also some EAs employed to address the cloud service optimization problems. Gao et al. \citeyearpar{num7} introduced the Enhanced Multi-Objective Jellyfish Search (EMOJS) algorithm to tackle the intricate cloud service optimization problem effectively. EMOJS was built upon MOJS \citep{num45}, but includes the introduction of four enhancement strategies to bolster its search capabilities. Furthermore, the team incorporated three enhancement strategies into the original Multi-Objective Gray Wolf Optimizer (MOGWO) algorithm \citep{num46}, resulting in the development of the strengthened MOGWO algorithm \citep{num47}, which proved to be efficient in solving the cloud service preference model.

Physical optimization algorithms draw inspiration from the principles of physics. These algorithms, characterized by high stochasticity, integrate global and local search methods to address the cloud service optimization problem effectively. Common examples of such algorithms include chaotic optimization algorithm, SA algorithms, and gravitational search algorithms \citep{num48}. For instance, Tao et al. \citeyearpar{num22} introduced an enhanced chaotic optimization algorithm by simulating nonlinear dynamical processes in chaotic systems. This algorithm employs adaptive chaos optimization and roulette wheel selection as operators. To leverage multicore computational resources efficiently and minimize the time consumption of adaptive chaos optimization, a parallel mechanism is implemented using full connectivity and reflection migration techniques. The resulting parallel adaptive chaos optimization based on reflection migration with full connectivity is specifically designed to address the cloud service optimization problem. SA emulate the annealing process of metallic materials, wherein the material is heated and then gradually cooled to enhance the metal structure. Wu et al. \citeyearpar{num49} proposed the use of an SA algorithm to optimize cloud service composition. They represent the metal cooling process as an iterative procedure, introducing a perturbation factor to update the solution set iteratively until reaching the final temperature and identifying the optimal solution.

A hybrid algorithm is a fusion of two or more heuristic optimization algorithms employed to address the cloud service optimization problem. The fundamental concept behind hybrid algorithms is to leverage the strengths of diverse algorithms and amalgamate them into a singular optimization algorithm, thereby enhancing its overall performance in terms of optimization metrics such as time or solution quality. For instance, Jin et al. \citeyearpar{num50} introduced a novel approach that combines the eagle strategy with the WOA. They employed uniform variation for global search to maintain the diversity of solutions. The WOA itself is an efficient optimization technique with high convergence speed \citep{num31}, enhancing the ability to escape from local optimal solutions by conducting local searches through an improved WOA for swift convergence to the optimal solution. Lastly, they proposed a novel hawk strategy using uniform variation and a modified WOA to address the cloud service optimization problem. In a similar, Wu et al. \citeyearpar{num51} enhanced the PSO algorithm by incorporating the outstanding global search capability of the SA algorithm. This was done to tackle the issue of the simple PSO algorithm potentially falling into local optimal solutions or experiencing premature convergence. They designed a hybrid PSO algorithm to address the challenge of cloud service optimization.

The NSGA-II algorithm, a prominent GA, has found widespread application in existing literature. The subsequent discussion delves into the mechanisms by which the NSGA-II algorithm addresses the challenge of cloud service optimization. The NSGA-II algorithm, as a representative multi-objective optimization algorithm, exhibits commendable operational speed and robustness, enabling the rapid derivation of resource allocation strategies  \citep{num44}. Employing a swift non-dominated sorting method, the algorithm organizes the population into strata using a crowding degree calculation  \citep{num52}, simplifying the overall algorithmic process. Furthermore, NSGA-II introduces a crowding index to assign fitness values to individuals at the same level post-sorting, along with a crowding mechanism to encourage uniform distribution along the Pareto front. Simultaneously, NSGA-II retains elite individuals, utilizing selected individuals and parent individuals to generate new populations, preserving dominant individuals and enhancing the evolutionary process. Consequently, it stands out as one of the most effective evolutionary multi-objective optimization algorithms to date. As illustrated in Figure 3, Algorithm \ref{alg2} delineates the process of the NSGA-II algorithm, addressing the challenge of cloud service optimization.


\begin{algorithm}[H]
	\caption{NAGA-II algorithm to solve IMfg-CSCOS problem}\label{alg2}
	Mathematical modeling of the IMfg-CSCOS problem; \\
	Creating Objective Functions and Constraint Functions; \\
	Manufacturing services in the service candidate set are coded using integer coding; \\
	Randomly generate an initial population $P_{Gen}$ with a total number of individuals $N$; \\
	Fast non-dominated sorting of the population $P_{Gen}$ according to the objective function and calculation of the crowding degree; \\
	\While{$Gen$\textless $MAX_{Gen}$}{
		Selection, crossover, and mutation operations are performed on individuals to produce a progeny population, $Q_{Gen}$, with the same number of individuals as the initial population; \\
		The parent and offspring populations were combined to obtain a combined population $P_{m}$ with 2$N$ individuals in the population;\\
		Fast non-dominated sorting of the population $P_{m}'s$ according to the objective function and calculation of the crowding degree;\\
		The fitness of individuals is calculated based on the non-dominated sorting rank and crowding, and the optimal $N$ individuals are retained by the elite retention strategy to generate a new parent population $P_{Gen+1}$;\\
		$Gen$=$Gen$+1;
		}
	Return the best solution
\end{algorithm}

\begin{figure}
	\centering
	\includegraphics[width=0.5\linewidth, height=0.65\textheight]{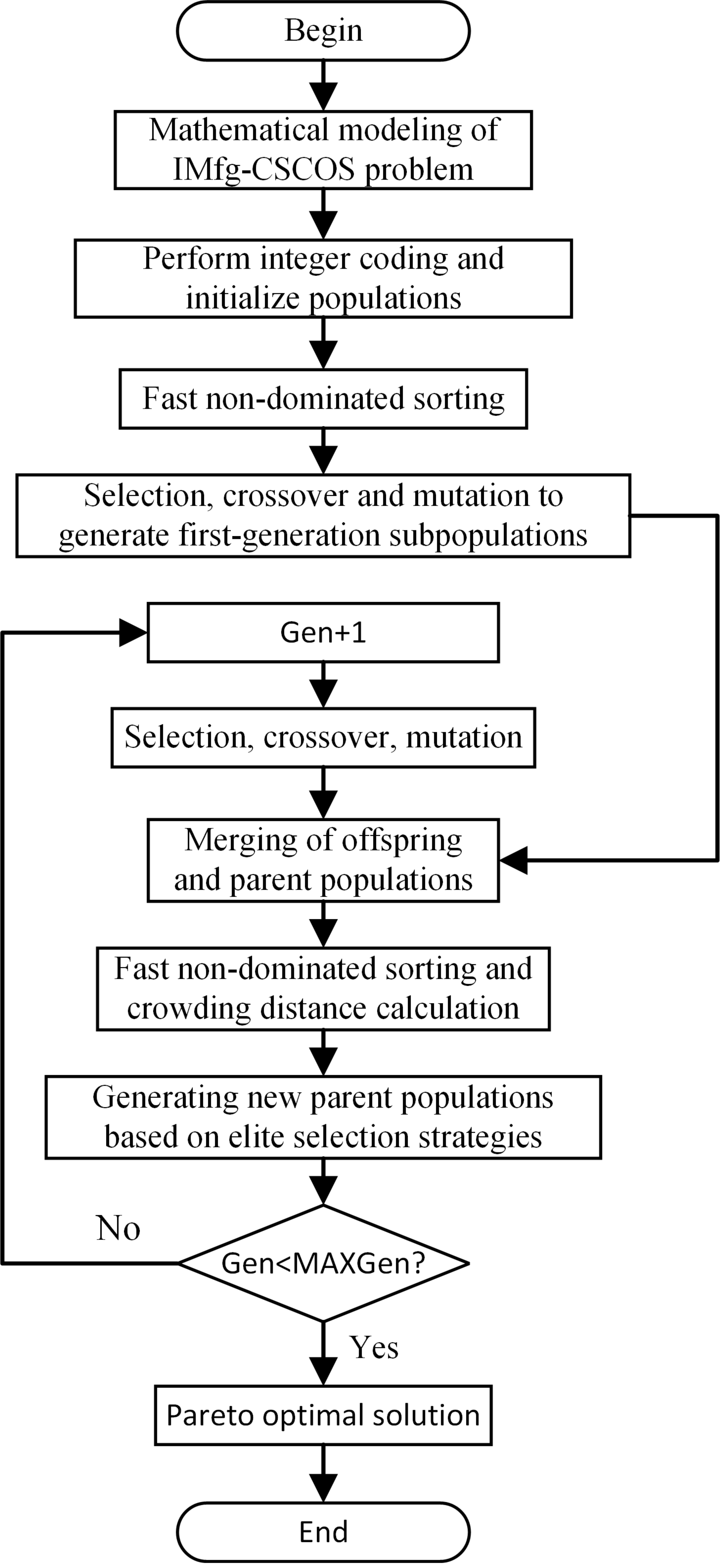}
	\caption{NSGA-II algorithm flow chart}
	\label{fig3}
\end{figure}


When addressing the cloud service optimization problem with heuristic algorithms, many commonly evaluation metrics are employed, including effectiveness, efficiency, distributivity, convergence, and other relevant factors. An algorithm’s effectiveness can be derived from the QoS value using an aggregation function. When comparing different algorithms, a higher QoS value in the generated solution set indicates that the algorithm’s solution set is more effective in addressing the problem. Additionally, accuracy and the number of non-dominated solutions to some extent also reflect the algorithm’s effectiveness. Efficiency is quantified by the algorithm’s execution time \citep{num53}, where more efficient algorithms can swiftly find solutions or approximations, conserving computational resources and time. A shorter execution time signifies a more efficient algorithm. The degree of distribution, or distributivity, assesses the distribution of the Pareto solution set generated by the heuristic algorithm in the target space, encompassing the homogeneity and diversity of the solution set. An exemplary algorithm should produce uniformly distributed solutions in a wide range of the objective space. In optimization problems, spacing and spread are commonly used metrics to gauge algorithmic distributivity \citep{num54}. Spacing represents the standard deviation of the minimum Euclidean distance between each solution and other solutions, where a smaller spacing value indicates a more uniformly distributed set of solutions. Meanwhile, spread measures the maximum Euclidean distance from each solution to other solutions \citep{num47}, serving as an indicator of the comprehensiveness of the obtained set of solutions. Convergence is an indicator of how closely the algorithm-generated solutions approach the true Pareto front, measured by calculating the distance between the solution set and the Pareto front. The convergence metric, often referred to as the generation distance (GD) \citep{num51}, quantifies the square of the sum of the Euclidean distances from the algorithm-obtained solution set to the nearest reference point on the true Pareto front. A smaller GD value signifies that the solutions generated by the algorithm are closer to the true Pareto front, indicating better algorithmic performance. In addition, there are comprehensive metrics for heuristic algorithms that assess both convergence and diversity. Hypervolume (HV) measures algorithm performance by calculating the volume between the generated solution and the true Pareto front \citep{num16}. A larger HV value signifies better convergence and diversity of the solutions generated by the algorithm. The inverted generation distance (IGD) is employed to evaluate an algorithm’s performance in an optimization problem \citep{num25}. It reflects the convergence and diversity of the solutions generated by the algorithm by measuring the distance between a set of non-dominated solutions generated by the algorithm and the true Pareto frontier. A smaller IGD value indicates that the algorithm produces a solution that is closer to the true frontier, demonstrating better solution quality.

In summary, while some progress has been made using heuristic algorithms to address IMfg-CSCOS, there are notable shortcomings. First, the design process of heuristic algorithms is intricate, and parameter tuning is challenging. Developing an effective heuristic algorithm requires a profound understanding of the problem, involving manual algorithm design and tuning of multiple parameters to adapt to specific issues. Second, heuristic algorithms may take an extended time to converge to the optimal solution, and there is a risk of getting trapped in local optima, impeding the global search for the optimal solution. Lastly, heuristic algorithms exhibit limited adaptability to the dynamic nature of the IMfg-CSCOS problem, making it challenging to cope with the swiftly changing manufacturing environment in intelligent manufacturing. In contrast, using RL for optimization does not necessitate explicit coding of the problem; instead, it autonomously learns the optimal policy based on rewards returned by the environment, eliminating the tedious manual coding process. Moreover, RL allows agents to adapt their strategies to the evolving manufacturing environment. Therefore, RL holds a distinct advantage in addressing the cloud service optimization problem.

\subsection{RL-based Optimization Algorithms}\label{subsec4.2}

\begin{figure}
	\centering
	\includegraphics[width=0.7\linewidth, height=0.37\textheight]{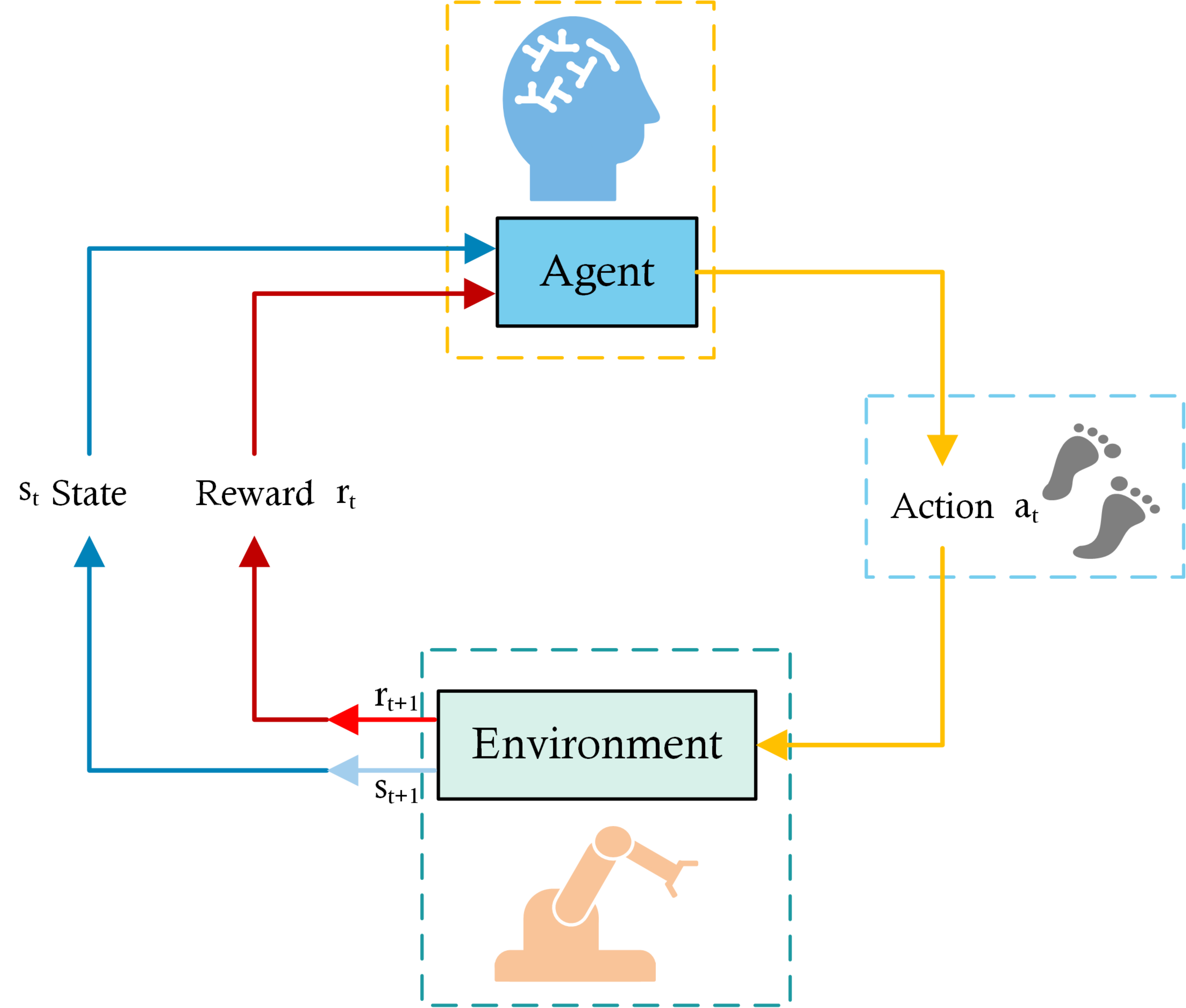}
	\caption{RL flow}
	\label{fig4}
\end{figure}

RL is a learning paradigm that establishes a mapping from environmental states to actions, with the objective of enabling the agent to accumulate the maximum reward throughout its interactions with the environment \citep{num55}. The foundational principle of RL centers on learning through a process of trial and error, which unfolds in two distinct steps. First, the agent engages with a specific environment, interacting to perceive outcomes such as receiving rewards or penalties. Subsequently, the agent assimilates the outcomes of its interactions with the environment, developing its optimization strategy. The objective of this strategy is to maximize rewards while minimizing penalties, analogous to the instincts of animals to avoid harm. Consequently, RL embodies a goal-oriented learning paradigm reliant on interactive experiences. As depicted in Figure \ref{fig4}, a typical RL model encompasses five primary elements: agent, environment, state, action, and reward. In the RL process, the agent evaluates the current environment and decides on an action (a $\in$ action) based on its strategy, leading to a transition from the current state ($S_{t}$) to the next state ($S_{t+1}$). Simultaneously, the agent receives an environmental reward ($R_{t}$). This iterative process persists until convergence.

RL incorporates the foundational structure of dynamic programming and Markov decision processes. It proves particularly valuable in addressing sequential optimization problems, such as cloud service optimization, offering the following advantages:

(i)The result derived from RL can be viewed as an optimal policy. The essence of a policy is an optimal rule, which can be mathematically conceptualized as a mapping, as illustrated in Equation (\ref{eq28}):

\begin{equation}\label{eq28}
\left. \pi(x): S\rightarrow A \right.
\end{equation}

The policy $\pi$ representation in Equation (\ref{eq28}) denotes a mapping from the finite state space to the action set. Consequently, specifying any state, \textit{S}, reveals the optimal action for the current state. This offers the advantage of effectively addressing various perturbations in the system. In the event of a perturbation, RL often adapts without requiring a complete re-optimization, in contrast to heuristic algorithms such as GA and PSO, which necessitate restarting the optimization process from the beginning.

(ii)RL facilitates the transfer of solutions across similar optimization problems, presenting a significant advantage in addressing the cloud service optimization problem. In cases where new problems exhibit strong correlations, a neural network–based RL algorithm, once applied to one, holds the potential to solve others. This is in contrast to heuristic algorithms like GA and PSO, where each new problem set requires a complete optimization algorithm run for each instance, even when some correlation exists among these problems.

RL can be classified into two main approaches: policy gradient–based and value function–based. Given the challenges traditional RL algorithms face in addressing high-dimensional problems, researchers have integrated the advantages of deep learning into RL. This integration empowers RL to tackle previously intractable issues, including those related to high-dimensional state and action spaces, and to extract input features more efficiently. In policy gradient–based RL, deep neural networks approximate the policy function, and the policy gradient method is utilized to determine the optimal policy, while In value function–based RL, deep neural networks are employed to approximate the value function. Consequently, RL becomes more adept at handling intricate tasks and environments.

(1)RL based on strategy gradients. The policy gradient method is a widely employed approach for policy optimization in RL, involving the updating of parameters by computing the gradient of the policy parameters concerning the total expected reward, ultimately converging to the optimal policy. In RL, a policy can be defined as a function, where a policy function establishes the probability of taking a specific action in a given state. The strategy gradient approach parameterizes the strategy and seeks optimal parameters to maximize the overall return. Consequently, when addressing the IMfg-CSCOS problem, a deep neural network can be employed to represent the policy model, and the policy gradient method can be applied to optimize the policy. When tackling IMfg-CSCOS problems, policy gradient–based algorithms are often more effective compared to using DQN and its enhanced variants. This preference for policy gradient–based algorithms arises from their capability to optimize the total expected reward of the policy directly. This approach leads to the attainment of an optimal solution in an end-to-end manner, eliminating the need for intermediate, intricate steps. Key RL methods based on policy gradient include Deep Deterministic Policy Gradient (DDPG), Trust Region Policy Optimization \citep{num56}, and Actor-Critic Algorithms \citep{num57}, along with their respective improved versions.

DDPG, deep deterministic policy gradient algorithm, was introduced by Lillicrap et al. \citeyearpar{num58}. It adapts the idea of DQN to the continuous action domain. In the realm of solving the IMfg-CSCOS problem, Liu et al. \citeyearpar{num59} were among the earliest contributors, proposing a DDPG-based methodology for cloud manufacturing service composition. This approach utilizes DDPG to address the cloud service optimization problem in dynamic intelligent manufacturing environments. DDPG can be viewed as an extension of DQN to the continuous action space, capitalizing on the strengths of both DQN and Actor-Critic for enhanced performance in continuous action spaces. In the Actor-Critic architecture, the actor network learns parameterized policies through a policy gradient algorithm, and the critic network learns value functions to assess state–action pairs obtained from the Q-learning algorithm. To overcome training instability, DDPG introduces a dual network structure. The online network is updated in real time through backpropagation during training, while the parameters of the target network are updated at regular intervals.

The Actor-Critic architecture, commonly employed in policy gradient–based approaches, enhances policy interpretability by segregating the policy from the evaluation network. This not only facilitates a deeper understanding of the strategy’s operational mechanisms but also enables targeted modifications to the strategy. Chen et al. \citeyearpar{num26} introduced AV-MPO, an in-policy maximum a posteriori policy optimization using Gated Transformer-XL (GTrXL). This algorithm’s structure, based on the Actor-Critic architecture, reduces the learning complexity of the Actor-Critic network through feature extraction and dimensionality reduction of states facilitated by the attention mechanism module.

The attention mechanism \citep{num60} has proved to be highly effective in extracting pertinent features depicting relationships within data. Consequently, numerous studies have integrated this mechanism into models to enhance algorithmic performance. Wang et al. \citeyearpar{num61} introduced the Scheduled Attention Model, which combines attention mechanisms with RL. This model incorporates a multi-head attention mechanism to capture interdependencies among service providers and between service providers and manufacturing tasks. Employing an encoder–decoder policy model, it utilizes two levels of attention to determine the probability of service selection. The model is further trained using the RL algorithm REINFORCE with a baseline \citep{num62}.

(2)\textit{RL based on value functions}. The value function is a fundamental concept in RL, encompassing classical algorithms such as temporal difference learning \citep{num63}, Q-learning \citep{num64}, and later developments like the Deep Q-network (DQN), introduced by Mnih et al. \citeyearpar{num65}, and various improved versions of DQN \citep{num66,num67}. All of these are classic algorithms in RL that rely on the value function. The value function is essentially an estimate of the expected future cumulative reward. In the decision-making process of RL, the agent assesses different action strategies, and the value function is an effective means of evaluating the long-term rewards derived from taking a specific action in a particular state. The DQN algorithm, as discussed further below, demonstrates outstanding learning performance. It enhances Q-learning by transforming the challenge of determining the Q-value of each action into the training process of a Q-network. Despite its efficacy, the use of neural networks to approximate the value function of actions in RL introduces convergence instability. To address this issue, DQN incorporates the experience replay mechanism and objective network:

(i)\textit{Experience Replay Mechanism}. This mechanism involves storing data acquired during the system’s exploration of the environment. Subsequently, random samples are drawn from this stored data to update the parameters of the deep neural network.

(ii)\textit{Objective Network}. In RL, when a nonlinear function is employed to approximate the Q-value function, updating the Q-value tends to experience oscillations and manifest unstable learning behavior. To address this challenge, an objective network is introduced.

The application of RL in the field of composition optimization has gained increasing popularity. It is not only applied to classical composition optimization problems such as web service composition and the traveling salesman problem, but is also finding relevance in the field of IMfg-CSCOS. Liang et al. \citeyearpar{num68} introduced PD-DQN, a fusion of a prioritized experience replay mechanism and Dueling DQN \citep{num69}. Dueling DQN represents an enhancement over DQN, breaking down the Q-value function into two components: the state value function and the advantage function. This decomposition enhances the treatment of the relationship between different actions and states. The framework incorporates an innovative prioritized experience utilization rule, assigning a distinctive weight to each experience in the process of updating the network parameters with experience. Subsequently, a weighted summation is conducted, and the network parameters are adjusted based on the outcomes of this weighted summation. This methodology tackles the intricacies of addressing the cloud service composition optimization problem.

\begin{figure}
	\centering
	\includegraphics[width=0.9\linewidth, height=0.3\textheight]{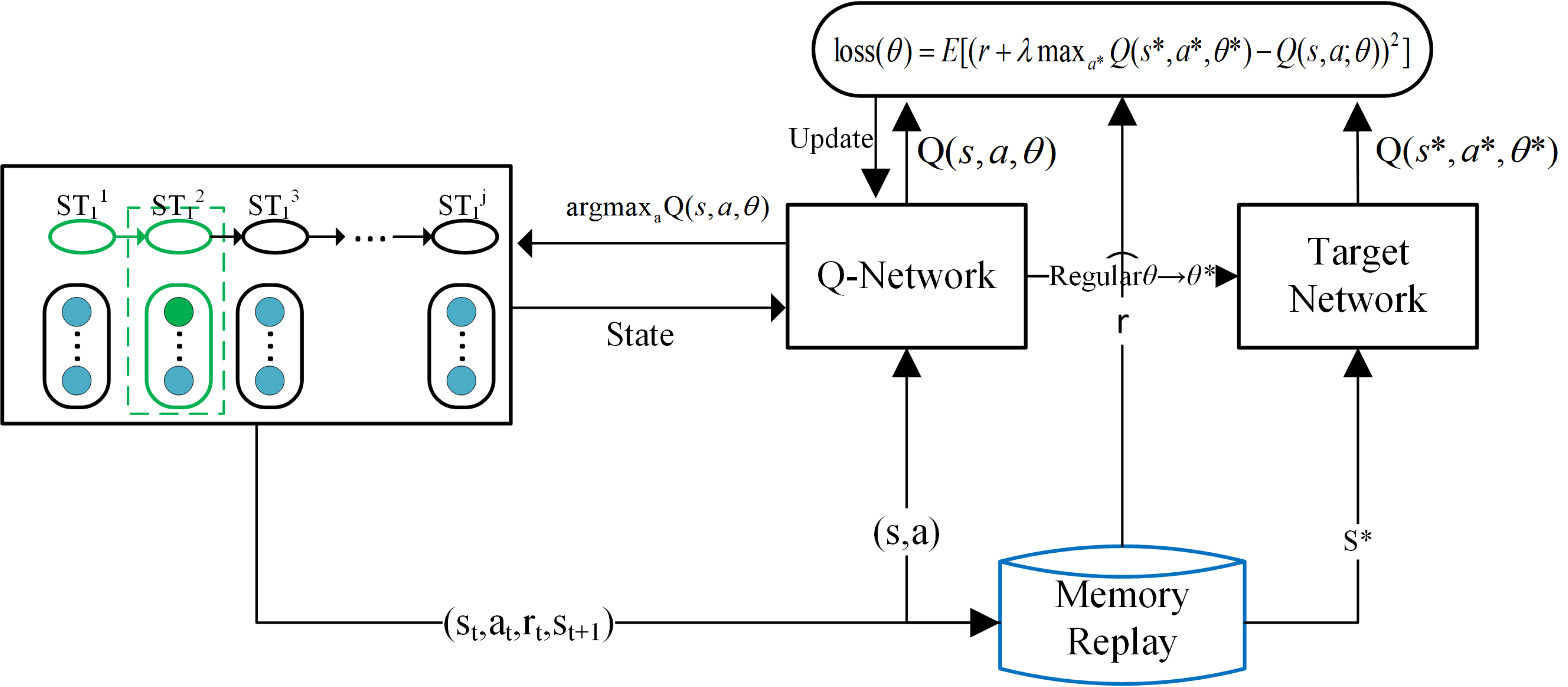}
	\caption{DQN algorithm flow chart}
	\label{fig5}
\end{figure}

The DQN algorithm stands out as one of the classical approaches in RL. In the DQN algorithm, a neural network is employed to learn the Q-value function, providing the algorithm with the capability to address high-dimensional state and action spaces, rendering it widely applicable. DQN incorporates a target network and an experience replay strategy, effectively balancing estimation error and variance, thereby enhancing algorithmic convergence. Illustrated in Figure \ref{fig5}, we use DQN with experience replay and a target network as an exemplar to elucidate in detail how RL tackles the IMfg-CSCOS problem. The procedural steps of DQN addressing the IMfg-CSCOS problem are delineated in Algorithm \ref{alg3}.



\begin{algorithm}[H]
	\caption{DQN algorithm to solve IMfg-CSCOS problem}\label{alg3}
	\KwIn{learning rate, minibatch size, sample size, discounted factor, initial exploration $\varepsilon$}
	Initialize Q-network and target-network parameters;\\
	\For{each episode}
	{
		\For{each each step t(subtask)}
		{
					With probability $\varepsilon$ select a random action(service) $a_{t}$, otherwise select $a_{t}=maxQ_{a}(s_{t},a_{t},\theta)$;\\
					Execute action $a_{t}$, observe reward $r_{t}$ and the next state(next subtask) $s_{t+1}$;\\
					Store experience ($s_{t}.a_{t},r_{t},s_{t+1}$) into the experience pool;\\
					Randomly sample minibatch of experience from the experience pool;\\
					Compute $Q^{*}$ using Target network:
					\begin{equation}
						Q^{*} = \left\{ \begin{matrix}
							{~~~~~~r_{t} ; ~~~~~~~~~~~~~~~~~~~~~~~~~~t~represents~the~final~time} \\
							{r_{t} + \lambda{max}_{a}Q\left( s_{t + 1},a_{t + 1},\theta^{*} \right) ; t~is~not~the~final~time}
						\end{matrix} \right.
					\end{equation}
		}
	}
	Return the best solution
\end{algorithm}

For RL algorithms, it is essential to employ evaluation metrics to gauge their performance. Standard metrics for evaluating RL algorithms encompass effectiveness, convergence, robustness, stability, scalability, and other pertinent factors. The effectiveness of an RL algorithm is primarily assessed through cumulative rewards, representing the total rewards accumulated by agents over a specified duration, where higher cumulative rewards signify superior agent performance. Convergence denotes the time or number of iterations required for the algorithm to attain a stable strategy. Liang et al. \citeyearpar{num68} gauged convergence by observing the number of iterations. Robustness measures the agent’s ability to withstand disturbances when the environment undergoes changes. Chen et al. \citeyearpar{num26} assessed robustness by introducing random faults in a cloud service setup. Stability reflects fluctuations made by the agent during the learning process, with smaller fluctuations indicating better stability. Scalability evaluates the agent’s adaptability to new environments or tasks, with a scalable algorithm capable of transitioning between tasks to enhance learning efficiency. Liu et al. \citeyearpar{num59} evaluated QoS performance across different service scales, where outstanding performance at diverse scales indicated increased algorithm scalability.

In summary, while the RL approach effectively addresses the challenges of complex and dynamic environments, demonstrating commendable scalability and robustness in solving cloud service optimization problems, it also encounters certain challenges and limitations. The decision-making process of RL algorithms is relatively intricate, yielding modeled decisions that may be challenging to interpret and comprehend, thereby potentially diminishing the interpretability of the system. Furthermore, the training process of RL introduces issues such as heightened complexity and unstable training. Finally, RL algorithms impose demanding requirements on both the quantity and quality of samples.

\subsection{Additional Optimization Algorithms}\label{subsec4.3}

In addition to the aforementioned optimization algorithms, numerous other noteworthy algorithms have been proposed by researchers to address the challenge of cloud service optimization. Within the cloud platform, a multitude of manufacturing services with comparable functionalities but varying service qualities exist. These services undergo comprehensive evaluation, and the most suitable services are recommended to customers while ensuring satisfaction of service demands. Liu et al. \citeyearpar{num8} introduced a similarity-enhanced hybrid group recommendation approach for cloud manufacturing. The algorithm initially categorizes users based on a similarity measure, followed by constructing a weighted sorted aggregation model to generate a recommendation list derived from representative users in each user group. Ma et al. \citeyearpar{num15} acknowledged that users possess distinct preferences for manufacturing service attributes. They extracted attribute weights from a vast dataset of historical records using variable-precision rough set theory mining. Subsequently, a combined ranking was determined by calculating the product of each service attribute value vector and the attribute weight vector. The Euclidean distance was then computed with the vector of service attribute values preferred by the user to obtain the combined ranking. Jiang et al. \citeyearpar{num18} proposed a combined weight assignment method that integrates hierarchical analysis and entropy value method for the comprehensive evaluation of cloud manufacturing services. By leveraging the acquired comprehensive evaluation values of manufacturing services, in conjunction with the trustworthiness of these services, a set of services that best aligns with user needs is recommended.

There exists a one-to-one or one-to-many matching relationship between manufacturing service providers and manufacturing demanders. This is due to the fact that in the cloud service optimization process, manufacturing tasks select manufacturing sub-services in the form of subtasks. In other words, a manufacturing subtask can call a manufacturing sub-service, or a manufacturing sub-service may be called by multiple manufacturing subtasks. Zhang et al. \citeyearpar{num70} developed a task-manufacturing service dynamic matching network model based on this matching relationship. The model continuously acquires and updates the states of manufacturing tasks and manufacturing services in real time, facilitating dynamic matching. Wang et al. \citeyearpar{num71} constructed a stable and satisfactory matching model for manufacturing services and manufacturing tasks. Intelligent manufacturing supply and demand parties serve as the main entities in this model, and the model addresses the problem through a bilateral matching method.

In the IMfg-CSCOS process, due to the correlations among metrics for each manufacturing service, some researchers have employed a grey correlation analysis to address the cloud service composition optimization problem. For example, Yuan et al. \citeyearpar{num11} assessed the efficacy of the current service composition by computing the grey correlation coefficient with respect to the worst or optimal composition solution. Indeed, mathematical methods have proved effective in tackling the cloud service optimization problem. For instance, Zhang et al. \citeyearpar{num72} employed augmented Lagrangian techniques to resolve the cloud service optimization problem. Hu et al. \citeyearpar{num73} addressed this optimization challenge by integrating tensor factorization (TF) and introduced a nonnegative TF-based predictor for acquiring the average attribute values of candidate services over a specified period. The predictor incorporates task, service, and temporal information into a low-dimensional space for low-rank approximation. Prediction-based data aid in optimizing time-dynamic solutions for the cloud service optimization problem through an enhanced multi-objective evolutionary optimizer, such as GrEA-X \citep{num73}.

\section{Research Hot Topics}\label{sec5}

To address the IMfg-CSCOS problem effectively, enhancing the service quality of manufacturing tasks requires a comprehensive and multiplicative understanding of the cloud service composition challenge. This involves emphasizing the synergies among diverse cloud service providers in the composition chain, comprehending the dynamic nature of the cloud service optimization environment in intelligent manufacturing, and exploring how RL can be better utilized to solve the cloud service optimization problem in the context of intelligent manufacturing.

(1) \textit{Collaboration ability among services}. Collaboration ability was initially introduced by Ansoff and Brandenburg\citeyearpar{num74}, who posited that strategic collaboration capabilities emerge when firms collaborate or possess complementary resources, enabling them to work together effectively. In the context of cloud services, the ability to collaborate is essential for achieving manufacturing tasks with greater efficiency. Additionally, there exists a level of collaboration ability among manufacturing service providers, often manifested in the form of compatibility and the sharing of resources between services. For instance, heightened efficiency in information exchange and smoother material transportation contribute to an increased level of collaboration ability among cloud manufacturing services. Conversely, a decline in the efficiency of information exchange and material transportation leads to a reduced level of collaboration ability among these services \citep{num75}. Hence, certain researchers have initiated investigations into the collaboration capabilities among services with the aim of enhancing customer satisfaction. Chen et al. \citeyearpar{num76} developed a composition synergy degree evaluation model to assess the synergistic impact among diverse services. Subsequently, they computed the composition synergy degree of the service composition chain using a synergy relationship matrix. Finally, the optimization of service composition took into account both the composition synergy degree and QoS. Xie et al. \citeyearpar{num39} introduced a two-phase methodology to tackle the cloud service optimization problem. They gauged service collaboration ability by quantifying the number of successful interactions a service provider has experienced in the past, incorporating a temporal dimension to evaluate the influence of each interaction on service collaboration ability. Ultimately, the integration of QoS and service collaboration ability serves to mitigate the probability of service failure. Hu et al. \citeyearpar{num77} developed a model for optimizing the composition of cloud manufacturing services, taking into account the ability of service collaboration. This model associates the ability of collaboration between services with factors such as migration cost, cooperation intensity, and cooperation intention. Ultimately, it significantly enhances the QoS in cloud service composition.

(2) \textit{The dynamic nature of the service composition environment}. In the context of intelligent manufacturing, the dynamics of manufacturing tasks and requirements may undergo changes, encompassing variations in the availability of services and the number of services required for each subtask. This dynamism represents a crucial feature of intelligent manufacturing, reflecting the real-time and evolving nature of the manufacturing environment. IMfg-CSCOS strategies must also adapt to such a dynamic environment, where swiftly generating relatively optimal solutions for cloud manufacturing in a dynamic cloud manufacturing environment holds greater significance, as it ensures a better alignment with user needs. Zhang et al. \citeyearpar{num70} considered the impact of an excessive load on the QoS of a service. Consequently, they developed a load queue model to assess the QoS dynamically. Ultimately, the dynamic QoS issue for service integration was addressed using a dynamic matching network approach with dynamic QoS and load as optimization objectives. Jiang et al. \citeyearpar{num43} proposed a variable-length coded GA designed for structurally changing incremental service composition to accommodate the variability and uncertainty of the dynamic environment in intelligent manufacturing. This approach utilizes variable-length coding schemes to depict the dynamics and uncertainties of the environment and addresses the problem by optimizing the process structure and the combination of service strengths. In dynamic manufacturing environments, not only does the QoS of a service change, but service anomalies may also occur during its execution. In this context, efficient reconfiguration of the original service composition under practical constraints becomes crucial. Wang et al. \citeyearpar{num78} proposed a dynamic service composition reconfiguration model for addressing service exceptions under practical constraints. The model considers the mandatory time constraints of the original service composition chain in the context of service anomalies, as well as practical constraints such as dynamic changes in the QoS. The solution to this model effectively addresses the issue of service anomalies occurring in the cloud service process. Wang et al. \citeyearpar{num79} further examined the coupled relationships of manufacturing resources, the selection of processing and transportation modes, and the reconfiguration adjustment of manufacturing resources. They constructed a dynamic service composition reconfiguration model that takes into account the actual constraints in real intelligent manufacturing. This model, combined with the proposed algorithm, effectively addressed the service anomaly problem.

(3) \textit{Cloud service optimization based on RL}. With the rapid advancement of AI technology, RL, as a subset of AI, has emerged as a prominent research focus for addressing cloud service optimization problems. At present, RL is applied in various domains, encompassing gaming \citep{num80}, autonomous driving \citep{num81,num82}, robot control \citep{num83,num84}, recommender systems \citep{num85}, the financial sector \citep{num86}, and job shop scheduling \citep{num87}. Therefore, RL presents a novel paradigm for addressing the challenges of IMfg-CSCOS. The utilization of RL in solving the cloud service optimization problem for intelligent manufacturing comes with notable advantages, attributed to the intricate and dynamic characteristics inherent in intelligent manufacturing. Through an analysis of the current research landscape, it is evident that there are a limited number of studies focusing on RL for IMfg-CSCOS, and most of them are recent. Primarily, the utilization of RL has proven to be more effective in addressing the challenges posed by dynamic environmental changes. Liang et al. \citeyearpar{num68} proposed an RL algorithm incorporating Dueling DQN with prioritized experience replay to address the issues in IMfg-CSCOS effectively. Liu et al. \citeyearpar{num59} employed a DDPG approach to address the cloud service composition problem in dynamic environments.  Second, neural networks facilitate the rapid extraction of relational features between service providers and manufacturing tasks. RL leverages these relational features to achieve more precise optimization of cloud services. Wang et al. \citeyearpar{num61} incorporated an attention mechanism to extract relational features between service providers and manufacturing tasks. Subsequently, they employed a combination of the encoder–decoder framework and RL to address the challenges of the IMfg-CSCOS problem.

\section{Application Scenarios}\label{sec6}

The application domain of IMfg-CSCOS is broad, mainly involving activities such as machining, manufacturing, designing, and the assembly of diverse mechanical components. The entire process of process manufacturing can be segmented into various manufacturing stages, each necessitating the provision of manufacturing services by a service provider. In the actual process of manufacturing, each segment involves numerous service providers offering similar functionalities but distinct nonfunctional attributes. Utilizing IMfg-CSCOS technology becomes imperative to discern and select the optimal cloud service that aligns with the user’s requirements from this pool of service providers. Moreover, it involves orchestrating a composition of these cloud services into a cohesive chain to accomplish the manufacturing task collaboratively. IMfg-CSCOS technology not only streamlines intricate and diverse manufacturing processes but also optimally fulfills customer manufacturing requirements during task completion. In the research domain of IMfg-CSCOS, its practical applications are diverse and extensive, spanning various manufacturing areas. This section elucidates some of these practical application scenarios.

Researchers have made numerous attempts to address the challenge of implementing IMfg-CSCOS technology in the domain of part machining. Component manufacturing is a pivotal aspect of the manufacturing industry, and IMfg-CSCOS technology can significantly contribute to this domain. The optimization of component manufacturing involves analyzing the material properties of the component, the machining process, and the overall production process. Intelligent manufacturing services can aid manufacturers in material selection, process optimization, and production planning. In addition, IMfg-CSCOS technology enables the development of distinct production processes tailored to different types of components, facilitating targeted optimization. Hence, IMfg-CSCOS technology holds significance in parts manufacturing. Several studies \citep{num40,num78} have illustrated this by applying IMfg-CSCOS technology to complete the hook-and-tail framework of a specific model in a forging plant, testing the practicability of IMfg-CSCOS technology. Wang et al. \citeyearpar{num88} utilized the production of large and complex cement equipment, such as cement mills, preheater systems, rotary kilns, and stacker reclaimers, as a research background, thereby testing IMfg-CSCOS technology in practical applications. Cai et al. \citeyearpar{num89} conducted an application case study using the production of a flight data recorder in an aerospace company.

IMfg-CSCOS technology also finds extensive applications in automotive manufacturing. First, during the automotive design stage, intelligent manufacturing services contribute to optimizing vehicle design and enhancing the overall performance for automakers. Second, in the course of automobile production, these services aid manufacturers in optimizing the production process, thereby improving production efficiency. Additionally, intelligent manufacturing services can assist automakers in optimizing supply chain management to ensure the timely delivery of parts. Therefore, IMfg-CSCOS technology has a broad range of applications in automotive manufacturing production. Liu et al. \citeyearpar{num59} provided an example of the production of automotive engine components, covering the manufacturing processes for exhaust gas recirculation channels, crankcases, valves, oil sumps, gearboxes, and connecting rods. Lim et al. \citeyearpar{num19} exemplified the processing of automotive fuel tank products, whereas Yuan et al. \citeyearpar{num90} conducted direct tests on the production of automotive parts as a practical application platform. Ren and Ren \citeyearpar{num91} utilized the machining process of a specific part in the intelligent workshops of Chery, Jaguar, and Land Rover automobiles as a prototype for a case application. In addition to the manufacturing of individual parts and components for automobiles, Zhu et al. \citeyearpar{num92} conducted a case study on the mass production of ordinary automobiles. This study primarily encompassed subtasks such as research and development, design, spare parts procurement, manufacturing, assembly, and sales. The algorithms were then integrated and tested for practical application.

In summary, IMfg-CSCOS technology holds significant potential across various domains, including parts manufacturing, automobile manufacturing, and beyond. As the manufacturing industry continues to evolve, the application scope of IMfg-CSCOS technology is expected to expand even further.

\section{Summary}\label{sec7}

In the context of the rapid advancement of modern manufacturing, IMfg-CSCOS technology is gaining increasing significance as a pivotal component in the manufacturing upgrade process. In the intelligent manufacturing platform, manufacturing resources are virtualized and deployed on the cloud platform as cloud services. The cloud platform undertakes centralized management of these manufacturing resources and optimizes the composition of cloud services based on task requirements submitted by users. 

In this paper, studies related to IMfg-CSCOS were reviewed and analyzed. After systematically reviewing and summarizing the relevant literature and research findings, a comprehensive examination and analysis of the current research status of IMfg-CSCOS was offered. Definitions of 11 commonly used indicators were derived through meticulous sorting and research. The article further delved into metrics-based approaches for cloud service optimization, categorizing them into heuristic-based optimization algorithms, RL-based optimization algorithms, and other algorithms. A summary of the advantages and disadvantages inherent in these distinct approaches was then provided. Heuristic-based optimization algorithms were further categorized into population algorithms, EAs, physical optimization algorithms, and hybrid algorithms. Meanwhile, RL-based optimization algorithms were shown to encompass two primary categories, namely, RL based on value functions and RL based on policy gradients. The definitions of these methods and related research were presented in detail. Subsequently, an in-depth analysis of the current research’s hot issues was conducted and, finally, a synopsis of the significance of cloud service optimization in practical application scenarios was provided.

In summary, IMfg-CSCOS stands out as a pivotal area in the ongoing development of the manufacturing industry. However, the current research results still require further exploration, primarily encompassing the following aspects:

(1) \textit{Insufficient consideration of the interests of both service providers and intelligent manufacturing platforms}. Current research predominantly focuses on cloud service optimization from the perspective of service demanders, overlooking the interests of service providers and intelligent manufacturing platforms. The first consideration is that service providers, functioning as significant manufacturing enterprises, must generate sufficient profit to sustain their survival. If prioritizing user interests results in an imbalance in the interests of service providers, significantly limiting the revenue potential for certain manufacturing service providers and demotivating them, it becomes necessary to consider the alignment of interests from the perspective of service providers. Second, operating the intelligent manufacturing platform necessitates resource consumption. Thus, minimizing this consumption becomes crucial for promoting the sustainable development of the platform.

(2) \textit{Lack of actual data and the absence of standardized datasets}. In real production environments, manufacturing enterprises frequently impose stringent confidentiality requirements on their production data and processes. Consequently, acquiring or sharing crucial manufacturing data becomes challenging. Moreover, due to the intricate nature of manufacturing operations, which involve diverse equipment and facilities, there are variations in data formats, protocols, and other aspects across different enterprises and equipment. The absence of a unified data standardization format complicates the collection and integration of data. This results in a significant expenditure linked to data acquisition and poses challenges in the collection process. Consequently, the QoS data utilized in the intelligent manufacturing cloud service composition rely heavily on randomly generated data from simulation experiments, lacking support from real datasets. This absence of credible experimental data undermines the reliability of the experiments.

(3) \textit{Absence of a comprehensive simulation experiment platform}. While significant progress has been made in IMfg-CSCOS research, there remains a gap in having a unified and standardized simulation experimental platform. First, this situation introduces challenges in comparing and validating the performance of various methods, as researchers conduct experiments in diverse simulation environments. Furthermore, simulation experiments in some studies might not sufficiently emulate the intricacies of real manufacturing environments. This deficiency in realism within simulation experiments may cast doubt on the applicability of research results in practical scenarios. Lastly, the diversity inherent in the manufacturing industry mandates the consideration of various types of manufacturing services and requirements. Existing simulation experiment platforms may lack the requisite diversity to encompass different manufacturing scenarios comprehensively. A well-designed simulation platform can offer more realistic scenarios, aiding researchers in validating their models and algorithms for real-world applications. The development of a standardized and widely adopted simulation experiment platform would enhance the sharing and reproducibility of research results.

(4) \textit{Theoretical analyses require refinement}. Given the complexity, dynamism, and diversity of real-world scenarios associated with IMfg-CSCOS problems, some research outcomes in IMfg-CSCOS may still exhibit shortcomings related to theoretical analysis. The absence of a comprehensive theoretical analysis may result in approaches that lean more towards empirical studies, lacking adequate interpretability when addressing IMfg-CSCOS issues. This complexity hinders a clear comprehension of the decision-making process of the methods, consequently limiting their practical applicability. A thorough theoretical analysis not only offers guidance for method optimization but also enhances the understanding of algorithm behavior across diverse scenarios.

\bmhead{Acknowledgments} 

This work was supported in part by the National Key Technologies Research and Development Program (2020YFB1712401, 2018YFB1701400), Key Research and Development Program in Henan Province(231111211900), Major Science and Technology Project in Henan Province (201300210500), Key Scientific Research Project of Colleges and Universities in Henan Province (23A520015), Henan Provincial Science and Technology Research Project (232102210090) . We thank International Science Editing (http://www.internationalscienceediting.com) for editing this manuscript.

\bmhead{Comflict of interests}
The authors declare that they have no competing interests.

\section{Declarations}\label{sec8}
\bmhead{Author contribution}
CL: Funding acquisition, Supervision, Methodology, Writing – review \& editing.
LL: Software, Visualization, Writing – original draft.
LS: Methodology,Conceptualization, Writing – review \& editing.

\bibliography{sn-bibliography}

\begin{thebibliography}{}
\providecommand{\doi}[1]{\url{https://doi.org/#1}}
\bibcommenthead

\bibitem[\protect\citeauthoryear{Akbaripour, Houshmand, van Woensel, and
  Mutlu}{Akbaripour et~al.}{2018}]{num14}
Akbaripour, H., Houshmand, M., van Woensel, T., \& Mutlu, N. (2018).
\newblock Cloud manufacturing service selection optimization and scheduling
  with transportation considerations: mixed-integer programming models.
\newblock {\em The International Journal of Advanced Manufacturing
  Technology\/}~95: 43--70.
\newblock \doi{10.1007/s00170-017-1167-3} .

\bibitem[\protect\citeauthoryear{Ansari, Yasmin, Naz, Zaffar, Ali, Moon, and
  Rho}{Ansari et~al.}{2022}]{num86}
Ansari, Y., Yasmin, S., Naz, S., Zaffar, H., Ali, Z., Moon, J., \& Rho, S.
  (2022).
\newblock A deep reinforcement learning-based decision support system for
  automated stock market trading.
\newblock {\em IEEE Access\/}~10: 127469--127501.
\newblock \doi{10.1109/ACCESS.2022.3226629} .

\bibitem[\protect\citeauthoryear{Ansoff and Brandenburg}{Ansoff and
  Brandenburg}{1967}]{num74}
Ansoff, H.I. \& Brandenburg, R.C. (1967).
\newblock A program of research in business planning.
\newblock {\em Management Science\/}~13: B219--B239.
\newblock \doi{10.1287/mnsc.13.6.B219} .

\bibitem[\protect\citeauthoryear{Back, Fogel, and Michalewicz}{Back
  et~al.}{1997}]{num41}
Back, T., Fogel, D.B., \& Michalewicz, Z. (1997).
\newblock {\em Handbook of Evolutionary Computation}.
\newblock GBR: IOP Publishing Ltd.

\bibitem[\protect\citeauthoryear{Beni and Wang}{Beni and Wang}{1993}]{num36}
Beni, G. \& Wang, J. (1993).
\newblock Swarm intelligence in cellular robotic systems.
\newblock In P.~Dario, G.~Sandini, and P.~Aebischer (Eds.), {\em Robots and
  Biological Systems: Towards a New Bionics?}, Berlin, Heidelberg, pp.\
  703--712. Springer Berlin Heidelberg.

\bibitem[\protect\citeauthoryear{Cai, Guo, Guo, Cai, and Xue}{Cai
  et~al.}{2019}]{num89}
Cai, A., Guo, Z., Guo, S., Cai, Y., \& Xue, X. (2019).
\newblock Optimization strategy of knowledge service composition in cloud
  manufacturing environment.
\newblock {\em Computer Integrated Manufacturing Systems\/}~25: 421--430.
\newblock \doi{10.13196/j.cims.2019.02.015} .

\bibitem[\protect\citeauthoryear{Cao, Wang, Kang, and Gao}{Cao
  et~al.}{2016}]{num37}
Cao, Y., Wang, S., Kang, L., \& Gao, Y. (2016).
\newblock A tqcs-based service selection and scheduling strategy in cloud
  manufacturing.
\newblock {\em The International Journal of Advanced Manufacturing
  Technology\/}~82: 235--251.
\newblock \doi{10.1007/s00170-015-7350-5} .

\bibitem[\protect\citeauthoryear{Chang, Yu, Hu, He, and Yu}{Chang
  et~al.}{2022}]{num87}
Chang, J., Yu, D., Hu, Y., He, W., \& Yu, H. (2022).
\newblock Deep reinforcement learning for dynamic flexible job shop scheduling
  with random job arrival.
\newblock {\em Processes\/}~10.
\newblock \doi{10.3390/pr10040760} .

\bibitem[\protect\citeauthoryear{Chen, Liu, Ling, and Wang}{Chen
  et~al.}{2019}]{num76}
Chen, Y., Liu, J., Ling, L., \& Wang, L. (2019).
\newblock Parallel manufacturing cloud service composition algorithm based on
  collaborative effect.
\newblock {\em Computer Integrated Manufacturing Systems\/}~25: 137--146.
\newblock \doi{10.13196/j.cims.2019.01.013} .

\bibitem[\protect\citeauthoryear{Chen, Wang, Liu, Zuo, and Niu}{Chen
  et~al.}{2019}]{num54}
Chen, Y., Wang, L., Liu, J., Zuo, L., \& Niu, Y. (2019).
\newblock Resource service composition optimization based on i-nsga-ii-jg
  algorithm for cloud manufacting.
\newblock {\em Computer Integrated Manufacturing Systems\/}~25: 2892--2904.
\newblock \doi{10.13196/j.cims.2019.11.018} .

\bibitem[\protect\citeauthoryear{Chen, Zhang, Wang, and Wang}{Chen
  et~al.}{2023}]{num26}
Chen, Z., Zhang, L., Wang, X., \& Wang, K. (2023).
\newblock Cloud–edge collaboration task scheduling in cloud manufacturing: An
  attention-based deep reinforcement learning approach.
\newblock {\em Computers $\&$ Industrial Engineering\/}~177: 109053.
\newblock \doi{https://doi.org/10.1016/j.cie.2023.109053} .

\bibitem[\protect\citeauthoryear{Chou and Truong}{Chou and
  Truong}{2020}]{num45}
Chou, J.S. \& Truong, D.N. (2020).
\newblock Multiobjective optimization inspired by behavior of jellyfish for
  solving structural design problems.
\newblock {\em Chaos, Solitons $\&$ Fractals\/}~135: 109738.
\newblock \doi{10.1016/j.chaos.2020.109738} .

\bibitem[\protect\citeauthoryear{Deb, Pratap, Agarwal, and Meyarivan}{Deb
  et~al.}{2002}]{num44}
Deb, K., Pratap, A., Agarwal, S., \& Meyarivan, T. (2002).
\newblock A fast and elitist multiobjective genetic algorithm: Nsga-ii.
\newblock {\em IEEE Transactions on Evolutionary Computation\/}~6: 182--197.
\newblock \doi{10.1109/4235.996017} .

\bibitem[\protect\citeauthoryear{Ding, Yan, Lei, and Xu}{Ding
  et~al.}{2019}]{num53}
Ding, T., Yan, G., Lei, Y., \& Xu, X. (2019).
\newblock A method of multi-level manufacturing service modeling and
  combinatorial optimal-selection.
\newblock {\em Journal of Beijing University of Aeronautics and
  Astronautics\/}~45: 1398--1405.
\newblock \doi{10.13700/j.bh.1001-5965.2018.0630} .

\bibitem[\protect\citeauthoryear{Dorigo}{Dorigo}{1992}]{num32}
Dorigo, M. (1992).
\newblock Optimization, learning and natural algorithms.

\bibitem[\protect\citeauthoryear{Eberhart and Kennedy}{Eberhart and
  Kennedy}{1995a}]{num30}
Eberhart, R. \& Kennedy, J. (1995)a.
\newblock A new optimizer using particle swarm theory.
\newblock In {\em MHS'95. Proceedings of the Sixth International Symposium on
  Micro Machine and Human Science}, Nagoya, Japan, pp.\  39--43. IEEE.

\bibitem[\protect\citeauthoryear{Eberhart and Kennedy}{Eberhart and
  Kennedy}{1995b}]{num38}
Eberhart, R. \& Kennedy, J. (1995)b.
\newblock A new optimizer using particle swarm theory.
\newblock In {\em MHS'95. Proceedings of the Sixth International Symposium on
  Micro Machine and Human Science}, Nagoya, Japan, pp.\  39--43. IEEE.

\bibitem[\protect\citeauthoryear{Einollah Jafarnejad~Ghomi and Qader}{Einollah
  Jafarnejad~Ghomi and Qader}{2019}]{num21}
Einollah Jafarnejad~Ghomi, A.M.R. \& Qader, N.N. (2019).
\newblock Service load balancing, task scheduling and transportation
  optimisation in cloud manufacturing by applying queuing system.
\newblock {\em Enterprise Information Systems\/}~13: 865--894.
\newblock \doi{10.1080/17517575.2019.1599448} .

\bibitem[\protect\citeauthoryear{Fu, Huang, Rao, Irissappane, Zhang, and Qu}{Fu
  et~al.}{2023}]{num85}
Fu, M., Huang, L., Rao, A., Irissappane, A.A., Zhang, J., \& Qu, H. (2023).
\newblock A deep reinforcement learning recommender system with multiple
  policies for recommendations.
\newblock {\em IEEE Transactions on Industrial Informatics\/}~19: 2049--2061.
\newblock \doi{10.1109/TII.2022.3209290} .

\bibitem[\protect\citeauthoryear{Gao, Yang, Wang, Fu, and Zhou}{Gao
  et~al.}{2023}]{num7}
Gao, Y., Yang, B., Wang, S., Fu, G., \& Zhou, P. (2023).
\newblock A multi-objective service composition method considering the
  interests of tri-stakeholders in cloud manufacturing based on an enhanced
  jellyfish search optimizer.
\newblock {\em Journal of Computational Science\/}~67: 101934.
\newblock \doi{https://doi.org/10.1016/j.jocs.2022.101934} .

\bibitem[\protect\citeauthoryear{Gao, Yang, Wang, Zhang, and Tang}{Gao
  et~al.}{2022}]{num47}
Gao, Y., Yang, B., Wang, S., Zhang, Z., \& Tang, X. (2022).
\newblock Bi-objective service composition and optimal selection for cloud
  manufacturing with qos and robustness criteria.
\newblock {\em Applied Soft Computing\/}~128: 109530.
\newblock \doi{10.1016/j.asoc.2022.109530} .

\bibitem[\protect\citeauthoryear{Gu, Holly, Lillicrap, and Levine}{Gu
  et~al.}{2017}]{num84}
Gu, S., Holly, E., Lillicrap, T., \& Levine, S. (2017).
\newblock Deep reinforcement learning for robotic manipulation with
  asynchronous off-policy updates.
\newblock In {\em 2017 IEEE International Conference on Robotics and Automation
  (ICRA)}, pp.\  3389–3396. IEEE Press.

\bibitem[\protect\citeauthoryear{Hasselt, Guez, and Silver}{Hasselt
  et~al.}{2016}]{num66}
Hasselt, H.v., Guez, A., \& Silver, D. (2016).
\newblock Deep reinforcement learning with double q-learning.
\newblock In {\em Proceedings of the Thirtieth AAAI Conference on Artificial
  Intelligence}, pp.\  2094–2100. AAAI Press.

\bibitem[\protect\citeauthoryear{Holland}{Holland}{1992}]{num29}
Holland, J.H. (1992).
\newblock {\em Adaptation in natural and artificial systems: an introductory
  analysis with applications to biology, control, and artificial intelligence}.
\newblock MIT press.

\bibitem[\protect\citeauthoryear{Houssein, Gad, Wazery, and Suganthan}{Houssein
  et~al.}{2021}]{num35}
Houssein, E.H., Gad, A.G., Wazery, Y.M., \& Suganthan, P.N. (2021).
\newblock Task scheduling in cloud computing based on meta-heuristics: Review,
  taxonomy, open challenges, and future trends.
\newblock {\em Swarm and Evolutionary Computation\/}~62: 100841.
\newblock \doi{https://doi.org/10.1016/j.swevo.2021.100841} .

\bibitem[\protect\citeauthoryear{Hu, Tian, Qi, Wu, and Liu}{Hu
  et~al.}{2023}]{num77}
Hu, Q., Tian, Y., Qi, H., Wu, P., \& Liu, Q. (2023).
\newblock Optimization method for cloud manufacturing service composition based
  on the improved artificial bee colony algorithm.
\newblock {\em Journal on Communications\/}~44: 200--210.
\newblock \doi{10.11959/j.issn.1000−436x.2023024} .

\bibitem[\protect\citeauthoryear{Hu, Wu, Yang, and Liu}{Hu
  et~al.}{2022}]{num73}
Hu, Y., Wu, F., Yang, Y., \& Liu, Y. (2022).
\newblock Tackling temporal-dynamic service composition in cloud manufacturing
  systems: A tensor factorization-based two-stage approach.
\newblock {\em Journal of Manufacturing Systems\/}~63: 593--608.
\newblock \doi{10.1016/j.jmsy.2022.05.008} .

\bibitem[\protect\citeauthoryear{Hu, Zhu, Zhang, Lui, and Wang}{Hu
  et~al.}{2019}]{num20}
Hu, Y., Zhu, F., Zhang, L., Lui, Y., \& Wang, Z. (2019).
\newblock Scheduling of manufacturers based on chaos optimization algorithm in
  cloud manufacturing.
\newblock {\em Robotics and Computer-Integrated Manufacturing\/}~58: 13--20.
\newblock \doi{https://doi.org/10.1016/j.rcim.2019.01.010} .

\bibitem[\protect\citeauthoryear{Jiang, Tang, Liu, and Zeng}{Jiang
  et~al.}{2022}]{num43}
Jiang, Y., Tang, L., Liu, H., \& Zeng, A. (2022).
\newblock A variable-length encoding genetic algorithm for incremental service
  composition in uncertain environments for cloud manufacturing.
\newblock {\em Applied Soft Computing\/}~123: 108902.
\newblock \doi{https://doi.org/10.1016/j.asoc.2022.108902} .

\bibitem[\protect\citeauthoryear{Jiang, Yi, and Zhu}{Jiang
  et~al.}{2020}]{num18}
Jiang, Z., Yi, D., \& Zhu, G. (2020).
\newblock Optimal selection of manufacturing cloud services considering
  elimination of fake cloud service.
\newblock {\em Computer Integrated Manufacturing Systems\/}~26: 2020--2029.
\newblock \doi{10.13196/j.cims.2020.08.002} .

\bibitem[\protect\citeauthoryear{Jin, Lv, Yang, and Liu}{Jin
  et~al.}{2022}]{num50}
Jin, H., Lv, S., Yang, Z., \& Liu, Y. (2022).
\newblock Eagle strategy using uniform mutation and modified whale optimization
  algorithm for qos-aware cloud service composition.
\newblock {\em Applied Soft Computing\/}~114: 108053.
\newblock \doi{https://doi.org/10.1016/j.asoc.2021.108053} .

\bibitem[\protect\citeauthoryear{Kirkpatrick}{Kirkpatrick}{1984}]{num34}
Kirkpatrick, S. (1984).
\newblock Optimization by simulated annealing: Quantitative studies.
\newblock {\em Journal of Statistical Physics\/}~34: 975--986.
\newblock \doi{https://doi.org/10.1007/BF01009452} .

\bibitem[\protect\citeauthoryear{Konda and Tsitsiklis}{Konda and
  Tsitsiklis}{1999}]{num57}
Konda, V. \& Tsitsiklis, J. (1999).
\newblock Actor-critic algorithms.
\newblock In S.~Solla, T.~Leen, and K.~M\"{u}ller (Eds.), {\em Advances in
  Neural Information Processing Systems}, Volume~12, pp.\  1008--1014. MIT
  Press.

\bibitem[\protect\citeauthoryear{Li, Chai, Hou, Zhang, Zhou, and Liu}{Li
  et~al.}{2018}]{num4}
Li, B., Chai, X., Hou, B., Zhang, L., Zhou, J., \& Liu, Y. (2018).
\newblock New generation artificial intelligence-driven intelligent
  manufacturing (ngaiim).
\newblock In {\em 2018 IEEE SmartWorld, Ubiquitous Intelligence $\&$ Computing,
  Advanced $\&$ Trusted Computing, Scalable Computing $\&$ Communications,
  Cloud $\&$ Big Data Computing, Internet of People and Smart City Innovation
  (SmartWorld/SCALCOM/UIC/ATC/CBDCom/IOP/SCI)}, pp.\  1864--1869. IEEE.

\bibitem[\protect\citeauthoryear{Li, Zhang, Wang, Tao, CAO, Jiang, Song, and
  Chai}{Li et~al.}{2010}]{num6}
Li, B., Zhang, L., Wang, S., Tao, F., CAO, J., Jiang, X., Song, X., \& Chai, X.
  (2010).
\newblock Cloud manufacturing: a new service-oriented networked manufacturing
  model.
\newblock {\em Computer integrated manufacturing system\/}~16: 0.
\newblock \doi{10.13196/j.cims.2010.01.3.libh.004} .

\bibitem[\protect\citeauthoryear{Li, Tao, Cheng, and Zhao}{Li
  et~al.}{2015}]{num5}
Li, J., Tao, F., Cheng, Y., \& Zhao, L. (2015).
\newblock Big data in product lifecycle management.
\newblock {\em The International Journal of Advanced Manufacturing
  Technology\/}~81: 667--684.
\newblock \doi{10.1007/s00170-015-7151-x} .

\bibitem[\protect\citeauthoryear{Li, Yao, and Liu}{Li et~al.}{2019}]{num75}
Li, Y., Yao, X., \& Liu, M. (2019).
\newblock Cloud manufacturing service composition optimization with improved
  genetic algorithm.
\newblock {\em Mathematical Problems in Engineering\/}~2019: 7194258.
\newblock \doi{10.1155/2019/7194258} .

\bibitem[\protect\citeauthoryear{Liang, Wen, Liu, Zhang, Zhang, and Wang}{Liang
  et~al.}{2021}]{num68}
Liang, H., Wen, X., Liu, Y., Zhang, H., Zhang, L., \& Wang, L. (2021).
\newblock Logistics-involved qos-aware service composition in cloud
  manufacturing with deep reinforcement learning.
\newblock {\em Robotics and Computer-Integrated Manufacturing\/}~67: 101991.
\newblock \doi{10.1016/j.rcim.2020.101991} .

\bibitem[\protect\citeauthoryear{Lillicrap, Hunt, Pritzel, Heess, Erez, Tassa,
  Silver, and Wierstra}{Lillicrap et~al.}{2015}]{num58}
Lillicrap, T.P., Hunt, J.J., Pritzel, A., Heess, N.M.O., Erez, T., Tassa, Y.,
  Silver, D., \& Wierstra, D. (2015).
\newblock Continuous control with deep reinforcement learning.
\newblock {\em CoRR\/}~abs/1509.02971: 1054--1054.
\newblock \doi{10.48550/arXiv.1509.02971} .

\bibitem[\protect\citeauthoryear{Lim, Xiong, and Wang}{Lim
  et~al.}{2022}]{num19}
Lim, M.K., Xiong, W., \& Wang, Y. (2022).
\newblock A three-tier programming model for service composition and optimal
  selection in cloud manufacturing.
\newblock {\em Computers $\&$ Industrial Engineering\/}~167: 108006.
\newblock \doi{https://doi.org/10.1016/j.cie.2022.108006} .

\bibitem[\protect\citeauthoryear{Liu, Chen, Liu, and Tekinerdogan}{Liu
  et~al.}{2023}]{num8}
Liu, J., Chen, Y., Liu, Q., \& Tekinerdogan, B. (2023).
\newblock A similarity-enhanced hybrid group recommendation approach in cloud
  manufacturing systems.
\newblock {\em Computers $\&$ Industrial Engineering\/}~178: 109128.
\newblock \doi{https://doi.org/10.1016/j.cie.2023.109128} .

\bibitem[\protect\citeauthoryear{Liu, Liang, Xiao, Zhang, Zhang, Zhang, and
  Wang}{Liu et~al.}{2022}]{num59}
Liu, Y., Liang, H., Xiao, Y., Zhang, H., Zhang, J., Zhang, L., \& Wang, L.
  (2022).
\newblock Logistics-involved service composition in a dynamic cloud
  manufacturing environment: A ddpg-based approach.
\newblock {\em Robotics and Computer-Integrated Manufacturing\/}~76: 102323.
\newblock \doi{10.1016/j.rcim.2022.102323} .

\bibitem[\protect\citeauthoryear{Liu, Song, Chu, Hou, and Peng}{Liu
  et~al.}{2017}]{num10}
Liu, Z., Song, C., Chu, D., Hou, Z., \& Peng, W. (2017).
\newblock An approach for multipath cloud manufacturing services dynamic
  composition.
\newblock {\em International Journal of Intelligent Systems\/}~32: 371--393.
\newblock \doi{https://doi.org/10.1002/int.21865} .

\bibitem[\protect\citeauthoryear{Lu and Cecil}{Lu and Cecil}{2016}]{num3}
Lu, Y. \& Cecil, J. (2016).
\newblock An internet of things (iot)-based collaborative framework for
  advanced manufacturing.
\newblock {\em The International Journal of Advanced Manufacturing
  Technology\/}~84: 1141--1152.
\newblock \doi{10.1007/s00170-015-7772-0} .

\bibitem[\protect\citeauthoryear{Lucic and Teodorovic}{Lucic and
  Teodorovic}{2002}]{num33}
Lucic, P. \& Teodorovic, D. (2002).
\newblock Transportation modeling: an artificial life approach.
\newblock In {\em 14th IEEE International Conference on Tools with Artificial
  Intelligence, 2002. (ICTAI 2002). Proceedings.}, Washington, DC, USA, pp.\
  216--223. IEEE.

\bibitem[\protect\citeauthoryear{Ma, Zhu, and Wang}{Ma et~al.}{2014}]{num15}
Ma, W., Zhu, L., \& Wang, W. (2014).
\newblock Cloud service selection model based on qos-aware in cloud
  manufacturing environment.
\newblock {\em Computer Integrated Manufacturing Systems\/}~20: 1246--1254.
\newblock \doi{10.13196/j.cims.2014.05.mawenlong.1246.9.20140528} .

\bibitem[\protect\citeauthoryear{Mirjalili and Lewis}{Mirjalili and
  Lewis}{2016}]{num31}
Mirjalili, S. \& Lewis, A. (2016).
\newblock The whale optimization algorithm.
\newblock {\em Advances in Engineering Software\/}~95: 51--67.
\newblock \doi{https://doi.org/10.1016/j.advengsoft.2016.01.008} .

\bibitem[\protect\citeauthoryear{Mirjalili, Saremi, Mirjalili, and dos
  S.~Coelho}{Mirjalili et~al.}{2016}]{num46}
Mirjalili, S., Saremi, S., Mirjalili, S.M., \& dos S.~Coelho, L. (2016).
\newblock Multi-objective grey wolf optimizer: A novel algorithm for
  multi-criterion optimization.
\newblock {\em Expert Systems with Applications\/}~47: 106--119.
\newblock \doi{10.1016/j.eswa.2015.10.039} .

\bibitem[\protect\citeauthoryear{Mnih, Kavukcuoglu, Silver, Graves, Antonoglou,
  Wierstra, and Riedmiller}{Mnih et~al.}{2013}]{num65}
Mnih, V., Kavukcuoglu, K., Silver, D., Graves, A., Antonoglou, I., Wierstra,
  D., \& Riedmiller, M. (2013).
\newblock Playing atari with deep reinforcement learning.
\newblock Preprint at \url{https://doi.org/10.48550/arXiv.1312.5602}.

\bibitem[\protect\citeauthoryear{Mnih, Kavukcuoglu, Silver, Rusu, Veness,
  Bellemare, Graves, Riedmiller, Fidjeland, Ostrovski, Petersen, Beattie,
  Sadik, Antonoglou, King, Kumaran, Wierstra, Legg, and Hassabis}{Mnih
  et~al.}{2015}]{num80}
Mnih, V., Kavukcuoglu, K., Silver, D., Rusu, A.A., Veness, J., Bellemare, M.G.,
  Graves, A., Riedmiller, M., Fidjeland, A.K., Ostrovski, G., Petersen, S.,
  Beattie, C., Sadik, A., Antonoglou, I., King, H., Kumaran, D., Wierstra, D.,
  Legg, S., \& Hassabis, D. (2015).
\newblock Human-level control through deep reinforcement learning.
\newblock {\em Nature\/}~518: 529--533.
\newblock \doi{10.1038/nature14236} .

\bibitem[\protect\citeauthoryear{Peters and Bagnell}{Peters and
  Bagnell}{2017}]{num62}
Peters, J. \& Bagnell, J.A. (2017).
\newblock {\em Policy Gradient Methods}, pp.\  982--985.
\newblock Boston, MA: Springer US.

\bibitem[\protect\citeauthoryear{Rashedi, Nezamabadi-pour, and
  Saryazdi}{Rashedi et~al.}{2009}]{num48}
Rashedi, E., Nezamabadi-pour, H., \& Saryazdi, S. (2009).
\newblock Gsa: A gravitational search algorithm.
\newblock {\em Information Sciences\/}~179: 2232--2248.
\newblock \doi{10.1016/j.ins.2009.03.004} .

\bibitem[\protect\citeauthoryear{Ren and Ren}{Ren and Ren}{2019}]{num91}
Ren, L. \& Ren, M. (2019).
\newblock Situation aware-adaptive decision-making mechanism of manufacturing
  composition service.
\newblock {\em Control and Decision\/}~34: 1277--1285.
\newblock \doi{10.13195/j.kzyjc.2017.1625} .

\bibitem[\protect\citeauthoryear{Sahoo and Lo}{Sahoo and Lo}{2022}]{num1}
Sahoo, S. \& Lo, C. (2022).
\newblock Smart manufacturing powered by recent technological advancements: A
  review.
\newblock {\em Journal of Manufacturing Systems\/}~64: 236--250.
\newblock \doi{10.1016/j.jmsy.2022.06.008} .

\bibitem[\protect\citeauthoryear{Sallab, Abdou, Perot, and Yogamani}{Sallab
  et~al.}{2017}]{num82}
Sallab, A.E., Abdou, M., Perot, E., \& Yogamani, S. (2017).
\newblock Deep reinforcement learning framework for autonomous driving.
\newblock {\em Electronic Imaging\/}~29: 70–76.
\newblock \doi{10.2352/issn.2470-1173.2017.19.avm-023} .

\bibitem[\protect\citeauthoryear{Schaul, Quan, Antonoglou, and Silver}{Schaul
  et~al.}{2016}]{num67}
Schaul, T., Quan, J., Antonoglou, I., \& Silver, D. (2016).
\newblock Prioritized experience replay.
\newblock Preprint at \url{https://doi.org/10.48550/arXiv.1511.05952}.

\bibitem[\protect\citeauthoryear{Schulman, Levine, Abbeel, Jordan, and
  Moritz}{Schulman et~al.}{2015}]{num56}
Schulman, J., Levine, S., Abbeel, P., Jordan, M., \& Moritz, P. (2015).
\newblock Trust region policy optimization.
\newblock In F.~Bach and D.~Blei (Eds.), {\em Proceedings of the 32nd
  International Conference on Machine Learning}, Volume~37, Lille, France, pp.\
   1889--1897. JMLR.org.

\bibitem[\protect\citeauthoryear{Srinivas and Deb}{Srinivas and
  Deb}{1994}]{num42}
Srinivas, N. \& Deb, K. (1994).
\newblock Muiltiobjective optimization using nondominated sorting in genetic
  algorithms.
\newblock {\em Evolutionary Computation\/}~2: 221--248.
\newblock \doi{10.1162/evco.1994.2.3.221} .

\bibitem[\protect\citeauthoryear{Sutton and Barto}{Sutton and
  Barto}{1998}]{num55}
Sutton, R. \& Barto, A. (1998).
\newblock Reinforcement learning: An introduction.
\newblock {\em IEEE Transactions on Neural Networks\/}~9: 1054--1054.
\newblock \doi{10.1109/TNN.1998.712192} .

\bibitem[\protect\citeauthoryear{Sutton}{Sutton}{1988}]{num63}
Sutton, R.S. (1988).
\newblock Learning to predict by the methods of temporal differences.
\newblock {\em Machine Learning\/}~3: 9--44.
\newblock \doi{10.1007/BF00115009} .

\bibitem[\protect\citeauthoryear{Tang, Zhao, Huang, and Zhang}{Tang
  et~al.}{2023}]{num17}
Tang, C., Zhao, S., Huang, T., \& Zhang, Q. (2023).
\newblock Cloud manufacturing service recommendation model based on qos
  multi-party heterogeneous evaluation and dual constraints of supply and
  demand.
\newblock {\em Computer Integrated Manufacturing Systems\/}~29: 2351--2362.
\newblock \doi{10.13196/j.cims.2023.07.018} .

\bibitem[\protect\citeauthoryear{Tao, LaiLi, Xu, and Zhang}{Tao
  et~al.}{2013}]{num22}
Tao, F., LaiLi, Y., Xu, L., \& Zhang, L. (2013).
\newblock Fc-paco-rm: A parallel method for service composition
  optimal-selection in cloud manufacturing system.
\newblock {\em IEEE Transactions on Industrial Informatics\/}~9: 2023--2033.
\newblock \doi{10.1109/TII.2012.2232936} .

\bibitem[\protect\citeauthoryear{Wang}{Wang}{2022}]{num71}
Wang, J. (2022).
\newblock {\em Research on Manufacturing Capacity Matching and Collaborative
  Plan Making under Cloud Manufacturing Mode}.
\newblock Ph.\ D. thesis, Wuhan University of Science and Technology.

\bibitem[\protect\citeauthoryear{Wang, Guo, Guo, Du, Li, and Wu}{Wang
  et~al.}{2018}]{num88}
Wang, L., Guo, C., Guo, S., Du, B., Li, X., \& Wu, R. (2018).
\newblock Rescheduling strategy of cloud service based on shuffled frog leading
  algorithm and nash equilibrium.
\newblock {\em The International Journal of Advanced Manufacturing
  Technology\/}~94: 3519--3535.
\newblock \doi{10.1007/s00170-017-1055-x} .

\bibitem[\protect\citeauthoryear{Wang, Zhang, Liu, and Zhang}{Wang
  et~al.}{2021}]{num25}
Wang, T., Zhang, P., Liu, J., \& Zhang, M. (2021).
\newblock Many-objective cloud manufacturing service selection and scheduling
  with an evolutionary algorithm based on adaptive environment selection
  strategy.
\newblock {\em Applied Soft Computing\/}~112: 107737.
\newblock \doi{https://doi.org/10.1016/j.asoc.2021.107737} .

\bibitem[\protect\citeauthoryear{Wang, Zhang, Liu, Zhao, and Wang}{Wang
  et~al.}{2022}]{num61}
Wang, X., Zhang, L., Liu, Y., Zhao, C., \& Wang, K. (2022).
\newblock Solving task scheduling problems in cloud manufacturing via attention
  mechanism and deep reinforcement learning.
\newblock {\em Journal of Manufacturing Systems\/}~65: 452--468.
\newblock \doi{https://doi.org/10.1016/j.jmsy.2022.08.013} .

\bibitem[\protect\citeauthoryear{Wang, Gao, Wang, and Zimmermann}{Wang
  et~al.}{2022}]{num40}
Wang, Y., Gao, S., Wang, S., \& Zimmermann, R. (2022).
\newblock An adaptive multiobjective multitask service composition approach
  considering practical constraints in fog manufacturing.
\newblock {\em IEEE Transactions on Industrial Informatics\/}~18: 6756--6766.
\newblock \doi{10.1109/TII.2021.3137831} .

\bibitem[\protect\citeauthoryear{Wang, Wang, Gao, Guo, and Yang}{Wang
  et~al.}{2021}]{num79}
Wang, Y., Wang, S., Gao, S., Guo, X., \& Yang, B. (2021).
\newblock Adaptive multi-objective service composition reconfiguration approach
  considering dynamic practical constraints in cloud manufacturing.
\newblock {\em Knowledge-Based Systems\/}~234: 107607.
\newblock \doi{10.1016/j.knosys.2021.107607} .

\bibitem[\protect\citeauthoryear{Wang, Wang, Kang, and Wang}{Wang
  et~al.}{2021}]{num78}
Wang, Y., Wang, S., Kang, L., \& Wang, S. (2021).
\newblock An effective dynamic service composition reconfiguration approach
  when service exceptions occur in real-life cloud manufacturing.
\newblock {\em Robotics and Computer-Integrated Manufacturing\/}~71: 102143.
\newblock \doi{10.1016/j.rcim.2021.102143} .

\bibitem[\protect\citeauthoryear{Wang, Schaul, Hessel, Van~Hasselt, Lanctot,
  and De~Freitas}{Wang et~al.}{2016}]{num69}
Wang, Z., Schaul, T., Hessel, M., Van~Hasselt, H., Lanctot, M., \& De~Freitas,
  N. (2016).
\newblock Dueling network architectures for deep reinforcement learning.
\newblock In {\em Proceedings of the 33rd International Conference on
  International Conference on Machine Learning - Volume 48}, pp.\  1995–2003.
  JMLR.org.

\bibitem[\protect\citeauthoryear{Watkins and Dayan}{Watkins and
  Dayan}{1992}]{num64}
Watkins, C.J.C.H. \& Dayan, P. (1992).
\newblock Q-learning.
\newblock {\em Machine Learning\/}~8: 279--292.
\newblock \doi{10.1007/BF00992698} .

\bibitem[\protect\citeauthoryear{Wei, Zhao, and Shu}{Wei et~al.}{2012}]{num24}
Wei, L., Zhao, Q., \& Shu, H. (2012).
\newblock Adaptive adjustment of composite cloud service based on qos for cloud
  manufacturing environment.
\newblock {\em Journal of Lanzhou University(Natural Sciences)\/}~48: 98--104.
\newblock \doi{10.13885/j.issn.0455-2059.2012.04.015} .

\bibitem[\protect\citeauthoryear{Wu, Greer, Rosen, and Schaefer}{Wu
  et~al.}{2013}]{num9}
Wu, D., Greer, M.J., Rosen, D.W., \& Schaefer, D. (2013).
\newblock {Cloud Manufacturing: Drivers, Current Status, and Future Trends}.
\newblock Volume Volume 2: Systems; Micro and Nano Technologies; Sustainable
  Manufacturing of {\em International Manufacturing Science and Engineering
  Conference}, Madison, Wisconsin, USA., pp.\  V002T02A003. Proceedings of the
  ASME 2013 International Manufacturing Science and Engineering Conference
  collocated with the 41st North American Manufacturing Research Conference.

\bibitem[\protect\citeauthoryear{Wu, Jia, and Cheng}{Wu et~al.}{2020}]{num51}
Wu, Y., Jia, G., \& Cheng, Y. (2020).
\newblock Cloud manufacturing service composition and optimal selection with
  sustainability considerations: a multi-objective integer bi-level
  multi-follower programming approach.
\newblock {\em International Journal of Production Research\/}~58: 6024--6042.
\newblock \doi{10.1080/00207543.2019.1665203} .

\bibitem[\protect\citeauthoryear{Wu, Jia, Luan, Wang, and Gao}{Wu
  et~al.}{2018}]{num49}
Wu, Y., Jia, G., Luan, S., Wang, L., \& Gao, Y. (2018).
\newblock Resource allocation model based on cloud manufacturing.
\newblock {\em Systems Engineering\/}~36: 122--128 .

\bibitem[\protect\citeauthoryear{Xie, Tan, Zheng, Zhao, Huang, and Sun}{Xie
  et~al.}{2021}]{num39}
Xie, N., Tan, W., Zheng, X., Zhao, L., Huang, L., \& Sun, Y. (2021).
\newblock An efficient two-phase approach for reliable collaboration-aware
  service composition in cloud manufacturing.
\newblock {\em Journal of Industrial Information Integration\/}~23: 100211.
\newblock \doi{https://doi.org/10.1016/j.jii.2021.100211} .

\bibitem[\protect\citeauthoryear{Xiong, Lim, Tseng, and Wang}{Xiong
  et~al.}{2023}]{num27}
Xiong, W., Lim, M.K., Tseng, M.L., \& Wang, Y. (2023).
\newblock An effective adaptive adjustment model of task scheduling and
  resource allocation based on multi-stakeholder interests in cloud
  manufacturing.
\newblock {\em Advanced Engineering Informatics\/}~56: 101937.
\newblock \doi{https://doi.org/10.1016/j.aei.2023.101937} .

\bibitem[\protect\citeauthoryear{Xiong, Wang, Zhang, and Li}{Xiong
  et~al.}{2016}]{num81}
Xiong, X., Wang, J., Zhang, F., \& Li, K. (2016).
\newblock Combining deep reinforcement learning and safety based control for
  autonomous driving .

\bibitem[\protect\citeauthoryear{Xu, Ba, Kiros, Cho, Courville, Salakhudinov,
  Zemel, and Bengio}{Xu et~al.}{2015}]{num60}
Xu, K., Ba, J., Kiros, R., Cho, K., Courville, A., Salakhudinov, R., Zemel, R.,
  \& Bengio, Y. (2015).
\newblock Show, attend and tell: Neural image caption generation with visual
  attention.
\newblock In F.~Bach and D.~Blei (Eds.), {\em Proceedings of the 32nd
  International Conference on Machine Learning}, Volume~37, Lille, France, pp.\
   2048--2057. PMLR.

\bibitem[\protect\citeauthoryear{Xu}{Xu}{2012}]{num2}
Xu, X. (2012).
\newblock From cloud computing to cloud manufacturing.
\newblock {\em Robotics and Computer-Integrated Manufacturing\/}~28: 75--86.
\newblock \doi{10.1016/j.rcim.2011.07.002} .

\bibitem[\protect\citeauthoryear{Yao, Xing, Zeng, and Wen}{Yao
  et~al.}{2021}]{num28}
Yao, J., Xing, B., Zeng, J., \& Wen, J. (2021).
\newblock Survey on cloud manufacturing service composition.
\newblock {\em Computer Science\/}~48: 245--255.
\newblock \doi{10.11896/jsjkx.200800173} .

\bibitem[\protect\citeauthoryear{Yuan, Cai, Zhou, Sun, Gu, and Huang}{Yuan
  et~al.}{2021}]{num90}
Yuan, M., Cai, X., Zhou, Z., Sun, C., Gu, W., \& Huang, J. (2021).
\newblock Dynamic service resources scheduling method in cloud manufacturing
  environment.
\newblock {\em International Journal of Production Research\/}~59: 542--559.
\newblock \doi{10.1080/00207543.2019.1697000} .

\bibitem[\protect\citeauthoryear{Yuan, Zhou, Cai, Sun, and Gu}{Yuan
  et~al.}{2020}]{num11}
Yuan, M., Zhou, Z., Cai, X., Sun, C., \& Gu, W. (2020).
\newblock Service composition model and method in cloud manufacturing.
\newblock {\em Robotics and Computer-Integrated Manufacturing\/}~61: 101840.
\newblock \doi{https://doi.org/10.1016/j.rcim.2019.101840} .

\bibitem[\protect\citeauthoryear{Zhang, Zhang, Xu, and Zhong}{Zhang
  et~al.}{2018}]{num72}
Zhang, G., Zhang, Y., Xu, X., \& Zhong, R.Y. (2018).
\newblock An augmented lagrangian coordination method for optimal allocation of
  cloud manufacturing services.
\newblock {\em Journal of Manufacturing Systems\/}~48: 122--133.
\newblock \doi{10.1016/j.jmsy.2017.11.008} .

\bibitem[\protect\citeauthoryear{Zhang, Geng, Bruce, Caluwaerts, Vespignani,
  SunSpiral, Abbeel, and Levine}{Zhang et~al.}{2017}]{num83}
Zhang, M., Geng, X., Bruce, J., Caluwaerts, K., Vespignani, M., SunSpiral, V.,
  Abbeel, P., \& Levine, S. (2017).
\newblock Deep reinforcement learning for tensegrity robot locomotion.
\newblock In {\em 2017 IEEE International Conference on Robotics and Automation
  (ICRA)}, Singapore, pp.\  634--641. IEEE.

\bibitem[\protect\citeauthoryear{Zhang and Zhao}{Zhang and Zhao}{2021}]{num52}
Zhang, Y. \& Zhao, X. (2021).
\newblock Manufacturing service composition recommendation based on bi-level
  programming model.
\newblock {\em J. Wuhan Univ.(Nat. Sci. Ed.)\/}~67: 547--554.
\newblock \doi{10.14188/j.1671-8836.2021.1010} .

\bibitem[\protect\citeauthoryear{Zhang, Zhang, Xu, Gao, and Xiao}{Zhang
  et~al.}{2018}]{num70}
Zhang, Z., Zhang, Y., Xu, X., Gao, F., \& Xiao, G. (2018).
\newblock Manufacturing service composition self-adaptive approach based on
  dynamic matching network.
\newblock {\em Journal of Software\/}~29: 3355--3373.
\newblock \doi{10.13328/j.cnki.jos.005475} .

\bibitem[\protect\citeauthoryear{Zhou, Gao, Yao, Zhang, Chan, and Lin}{Zhou
  et~al.}{2020}]{num16}
Zhou, J., Gao, L., Yao, X., Zhang, C., Chan, F.T., \& Lin, Y. (2020).
\newblock An adaptive dual-population evolutionary paradigm with adversarial
  search: Case study on many-objective service consolidation.
\newblock {\em Applied Soft Computing\/}~90: 106160.
\newblock \doi{https://doi.org/10.1016/j.asoc.2020.106160} .

\bibitem[\protect\citeauthoryear{Zhou and Yao}{Zhou and Yao}{2017a}]{num12}
Zhou, J. \& Yao, X. (2017)a.
\newblock De-caabc: differential evolution enhanced context-aware artificial
  bee colony algorithm for service composition and optimal selection in cloud
  manufacturing.
\newblock {\em The International Journal of Advanced Manufacturing
  Technology\/}~90: 1085--1103.
\newblock \doi{10.1007/s00170-016-9455-x} .

\bibitem[\protect\citeauthoryear{Zhou and Yao}{Zhou and Yao}{2017b}]{num13}
Zhou, J. \& Yao, X. (2017)b.
\newblock Hybrid teaching--learning-based optimization of correlation-aware
  service composition in cloud manufacturing.
\newblock {\em The International Journal of Advanced Manufacturing
  Technology\/}~91: 3515--3533.
\newblock \doi{10.1007/s00170-017-0008-8} .

\bibitem[\protect\citeauthoryear{Zhu, Li, Tang, and Sun}{Zhu
  et~al.}{2020}]{num23}
Zhu, H., Li, M., Tang, Y., \& Sun, Y. (2020).
\newblock A deep-reinforcement-learning-based optimization approach for
  real-time scheduling in cloud manufacturing.
\newblock {\em IEEE Access\/}~8: 9987--9997.
\newblock \doi{10.1109/ACCESS.2020.2964955} .

\bibitem[\protect\citeauthoryear{Zhu, Li, and Zhou}{Zhu et~al.}{2019}]{num92}
Zhu, L., Li, P., \& Zhou, X. (2019).
\newblock Ihdetbo: A novel optimization method of multi-batch subtasks
  parallel-hybrid execution cloud service composition for cloud manufacturing.
\newblock {\em Complexity\/}~2019: 7438710.
\newblock \doi{10.1155/2019/7438710} .

\end{thebibliography}


\end{document}